\documentclass[letterpaper]{article} 
\usepackage{aaai2026}  
\usepackage{times}  
\usepackage{helvet}  
\usepackage{courier}  
\usepackage[hyphens]{url}  
\usepackage{graphicx} 
\usepackage{natbib}  
\usepackage{caption} 
\usepackage{algorithm}
\usepackage{algorithmic}

\usepackage{newfloat}
\usepackage{listings}
\DeclareCaptionStyle{ruled}{labelfont=normalfont,labelsep=colon,strut=off} 
\floatstyle{ruled}
\newfloat{listing}{tb}{lst}{}
\floatname{listing}{Listing}
\usepackage{amsmath}
\usepackage{amssymb}
\usepackage{booktabs}
\usepackage{multirow}
\usepackage{makecell}
\usepackage{tcolorbox}
\usepackage{xcolor}

\title{Harnessing Vision-Language Models for Time Series Anomaly Detection}
\author{
    Zelin He\textsuperscript{\rm 1},
    Sarah Alnegheimish\textsuperscript{\rm 2},
    Matthew Reimherr\textsuperscript{\rm 1,3}\thanks{Work unrelated to Amazon. \\ {Code available at {https://github.com/ZLHe0/VLM4TS.}}}
}
\affiliations{
    \textsuperscript{\rm 1}Pennsylvania State University\\
    \textsuperscript{\rm 2}Massachusetts Institute of Technology\\
    \textsuperscript{\rm 3}Amazon\\
    zbh5185@psu.edu, smish@mit.edu, mlr36@psu.edu
}

\usepackage{bibentry}

\begin{document}

\maketitle

\begin{abstract}
Time-series anomaly detection (TSAD) has played a vital role in a variety of fields, including healthcare, finance, and sensor-based condition monitoring. Prior methods, which mainly focus on training domain-specific models on numerical data, lack the visual–temporal understanding capacity that human experts have to identify contextual anomalies. To fill this gap, we explore a solution based on vision language models (VLMs). Recent studies have shown the ability of VLMs for visual understanding tasks, yet their direct application to time series has fallen short on both accuracy and efficiency. To harness the power of VLMs for TSAD, we propose a two-stage solution, with (1) ViT4TS, a vision-screening stage built on a relatively lightweight pre-trained vision encoder, which leverages 2-D time series representations to accurately localize candidate anomalies; (2) VLM4TS, a VLM-based stage that integrates global temporal context and VLM's visual understanding capacity to refine the detection upon the candidates provided by ViT4TS. We show that without any time-series training, VLM4TS outperforms time-series pre-trained and from-scratch baselines in most cases, yielding a 24.6\% improvement in F1-max score over the best baseline. Moreover, VLM4TS also consistently outperforms existing language model-based TSAD methods and is on average 36 × more efficient in token usage. 
\end{abstract}

\section{Introduction}
\label{sec:intro}
Time series anomaly detection (TSAD) is an important task for maintaining safety and efficiency in many domains, such as cloud computing, industrial monitoring, and web services \cite{lavin2015evaluating}. One critical challenge is that time series signals usually exhibit diverse temporal scales and dynamic behaviors, demanding deep temporal understanding to distinguish true anomalies from benign fluctuations. For example, a sudden increase in spacecraft telemetry readings may be benign if it has recurred frequently in historical records, whereas a gradual drift that deviates from an established trend could be an anomaly \cite{hundman2018lstmdt}. However, most existing TSAD models are built based on domain-specific assumptions and are trained on numerical data, limiting them to detecting  surface-level anomalies without rich visual-temporal understanding capacities that human experts have \cite{kong2025position}.

\begin{figure}
  \centering
\includegraphics[width=\linewidth]{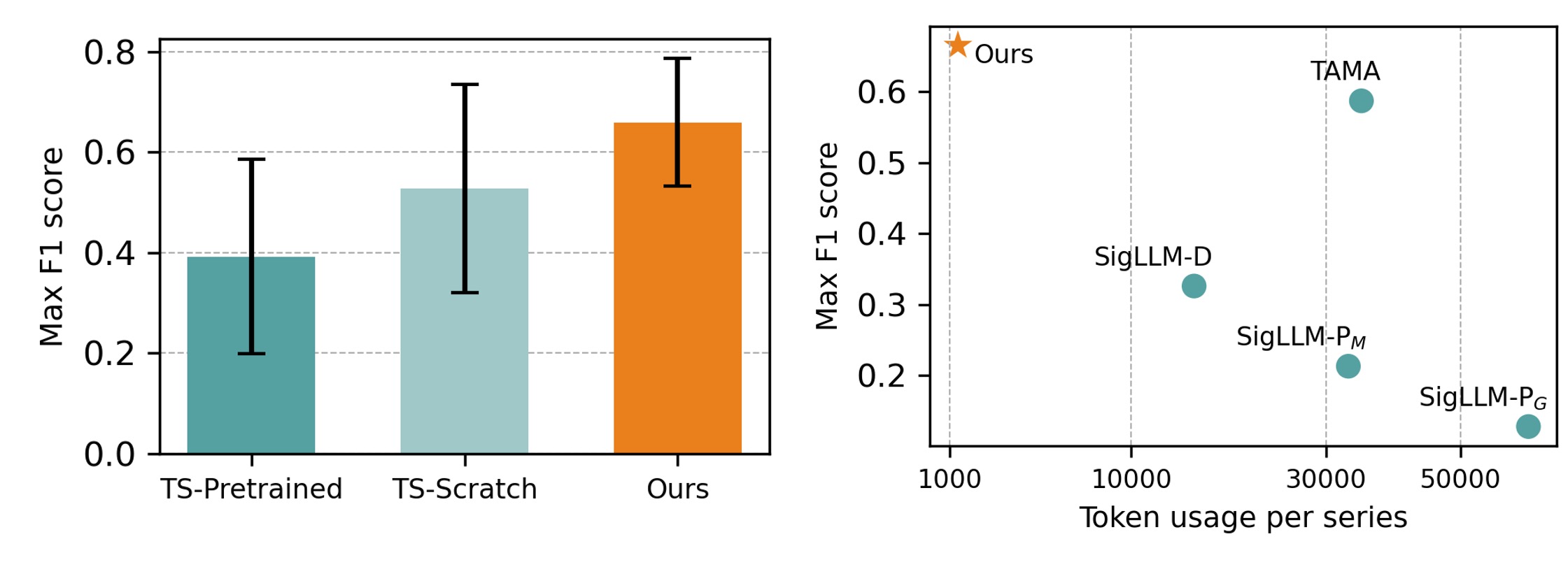}
  \caption{(Left) max F1 score averaged over all benchmarks, comparing VLM4TS to the best time-series-pretrained and from-scratch baselines. Error bars indicate standard deviation across datasets. (Right) max F1 score and token usage of VLM4TS versus language-model-based baselines.}
  \label{fig:introduction}
\end{figure}

\begin{figure}[t]
    \centering
    \includegraphics[width=0.85\linewidth]{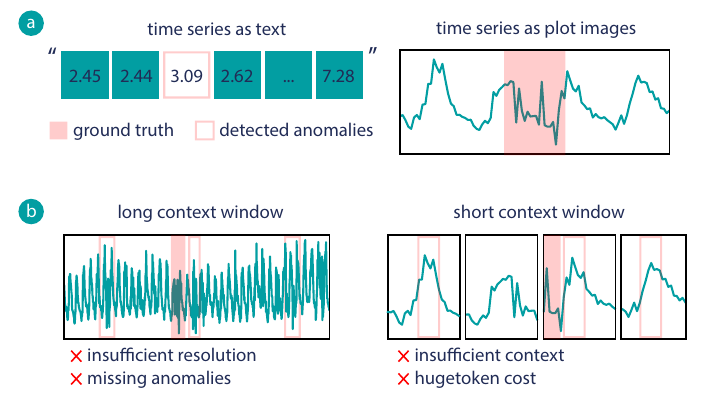}
    \caption{(a) Presenting time series as textual input for LLMs can obscure real anomalies and increase incorrect detection, whereas visualizing time series as line plots makes contextual anomalies, such as distortions, readily apparent. (b) The resolution–context dilemma in VLM-based TSAD: a global plot preserves long-range context but compresses details for detection (left), while small-window views maintain high resolution but provide limited context and incur high token cost (right).}
    \label{fig:motivation}
\end{figure}

Recent breakthroughs in multimodal foundation models have demonstrated human-level understanding across multiple modalities such as images, audio, and video \cite{wu2023multimodal}. However, there is not yet a well-developed time-series-language model that brings human-like understanding capacities for time series analysis and anomaly detection, largely because time series–text corpora are scarce \cite{kong2025position}. Vision–language models (VLMs) have become a good alternative for two main reasons. First, their vision transformer architecture mimics a human-expert visual inspection mechanism—comparing and contrasting patterns at multiple scales to localize and characterize anomalies. Compared with LLM-based approaches \cite{alnegheimish2024can}, a vision-based mechanism is especially effective at examining contextual anomalies that don't include extreme values but show deviation from normal signal patterns, see Figure \ref{fig:motivation}-(a). Second, pretraining on massive image–text datasets equips VLMs with strong inference and reasoning capabilities that generalize across domains \cite{nagar2024zero}. By encoding time series as visual representations, we can leverage VLMs’ rich visual–textual understanding to perform TSAD tasks. 

Recently, few pioneering works have explored VLM-based TSAD by rendering raw time series as line graphs with x-axis tick marks, then prompting the model to return anomaly intervals by referencing those tick labels \cite{zhuang2024see,zhou2024can,xu2025can}. In practice, however, such a naive application of VLMs faces a major resolution-context dilemma when selecting the window size, as demonstrated in Figure \ref{fig:motivation}-(b). With short context windows, we face challenges with \textbf{high cost, latency, and limited context}. To not only keep tick marks legible but also satisfy VLM input constraints, long series must be split into many small rolling window plots. For example, a 1,000-step sequence might produce around 20 images, consuming around 20,000 tokens and incurring high latency, which is untenable for real time monitoring across thousands of sensors \cite{alnegheimish2025m}. Furthermore, each plot captures only a brief temporal segment, thus preventing understanding over global trends or long range dependencies that distinguish true anomalies from transient noise. On the other hand, with long context windows, we face challenges with \textbf{limited resolution, poor localization, and missed anomalies}. Plotting long series as one single image reduces token usage but significantly degrades the image resolution. The VLM cannot pinpoint the exact interval boundaries via crowded x-axis ticks, even if an anomalous behavior is identified. Moreover, the volume of visual content overwhelms the attention of the model, causing it to overlook some of the anomalies.

In this paper, our goal is to harness the power of VLMs for TSAD by addressing the aforementioned resolution–context dilemma in a cost-effective manner. We propose a novel framework to decompose the anomaly detection task into two components: localization and verification.  The first step localizes anomalies via a lightweight pretrained vision encoder (ViT4TS), while the second step verifies anomalies with a heavier yet more powerful VLM (VLM4TS). We emphasize that both stages of our framework work in a zero‐shot setting—that is, neither ViT4TS nor VLM4TS is fine‐tuned on the in-domain time series data. ViT4TS uses off-the-shelf, image-pretrained weights to screen windows purely via visual pattern matching, and VLM4TS relies on the pretrained cross-modal understanding of a VLM, showing great generalizability to a broad range of TSAD tasks. Figure \ref{fig:introduction} showcases the gain in detection performance and token efficiency of our approach. To our knowledge, this is the first VLM-based TSAD solution to achieve both superior detection accuracy and practical computational and token efficiency. To summarize, our main contributions are listed as follows.

$\bullet$ Motivated by human diagnostic workflows, we explore a solution that casts a 1-D anomaly detection as a 2-D visual understanding problem with VLMs. To resolve the resolution–context dilemma in VLM application, we explore decomposing the problem into sequential localization and verification stages.  

$\bullet$ For the localization task, we propose ViT4TS, which leverages rich multi-scale cross-patch comparison to accurately localize anomaly candidates; then, for the verification task, we propose VLM4TS, that produce a final detection by filtering and refining on the detected anomalous intervals raised by ViT4TS with a deep temporal understanding on global temporal context.

$\bullet$ We demonstrate that our approach generalizes across domains and achieves state-of-the-art performance on multiple benchmark datasets without dataset-specific tuning, and requiring much less token usage than existing language model-based methods.

\section{Related Work}

\textbf{In-domain TSAD.} 
Early in-domain TSAD methods relied on statistical and distance-based techniques—e.g.,  threshold detectors and ARIMA residual analysis \cite{pena2013arima}, which demand clean, uncontaminated training data and extensive domain expertise. Later on, many deep learning-based methods were introduced, applying structures like RNNs and sequence autoencoders to detect anomalies via prediction or reconstruction error \cite{hundman2018lstmdt,malhotra2016lstmae,wong2022aer}. Some works also explored VAE- and GAN-based frameworks to model normal behavior probabilistically \cite{geiger2020tadgan,li2020anomaly,kieu2019outlier,yin2020anomaly}. More recently, Transformer-based models have shown the capacity to capture more complex temporal patterns for detection \cite{xu2021anomaly,tuli2022tranad}. For a comprehensive survey, see \cite{zamanzadeh2024deep}. Although highly effective in-domain, these approaches require large training sets and may not generalize well beyond the conditions they were trained on.

\textbf{Foundation Model-based and LLM-based TSAD.} 
Recent efforts pretrain general-purpose time series encoders \cite{gao2024units,zhang2024unimts} or build large forecasting foundation models \cite{rasul2023lag,ansari2024chronos,das2024decoder,goswami2024moment,shi2024time} that can be adapted for anomaly detection. However, these models usually excel at forecasting the next few time steps or encoding a short time window, but lack the capacity to perform human-like understanding over long-range temporal context, making them inefficient for detecting contextual anomalies across extended horizons. Prompting LLMs directly on numerical sequences has also been explored \cite{alnegheimish2024can}, yet they have shown that naive prompting underperforms some of the classical TSAD methods and comes with huge token costs.

\textbf{Vision and VLM-based TSAD.} 
Treating time series as images has shown promise across a range of tasks \cite{ni2025harnessing}, including classification \cite{costa2024fusion,kaewrakmuk2024multi,li2023time} and forecasting \cite{yang2023your,zeng2023pixels,semenoglou2023image}. However, the application in unsupervised anomaly detection remains relatively underexplored. There are few works exploring vision-based TSAD approaches using spectrograms or RP transformations \cite{namura2024training,lin2024hierarchical}, yet these approaches works purely with the vision modality. As a result, these pure vision-based approaches do not possess the multimodal reasoning and temporal understanding capabilities. Recent prototypes like TAMA \cite{zhuang2024see} and related methods \cite{zhou2024can,xu2025can} prompt VLMs on small rolling-window plots to assign anomaly scores, but they incur prohibitive token costs and offer only limited temporal context. Consequently, they cannot scale to real-world tasks or capture contextual anomalies based on long-range temporal understanding required for robust anomaly detection.

\section{Methodology}
\begin{figure*}
    \centering
    \includegraphics[width=0.85\linewidth]{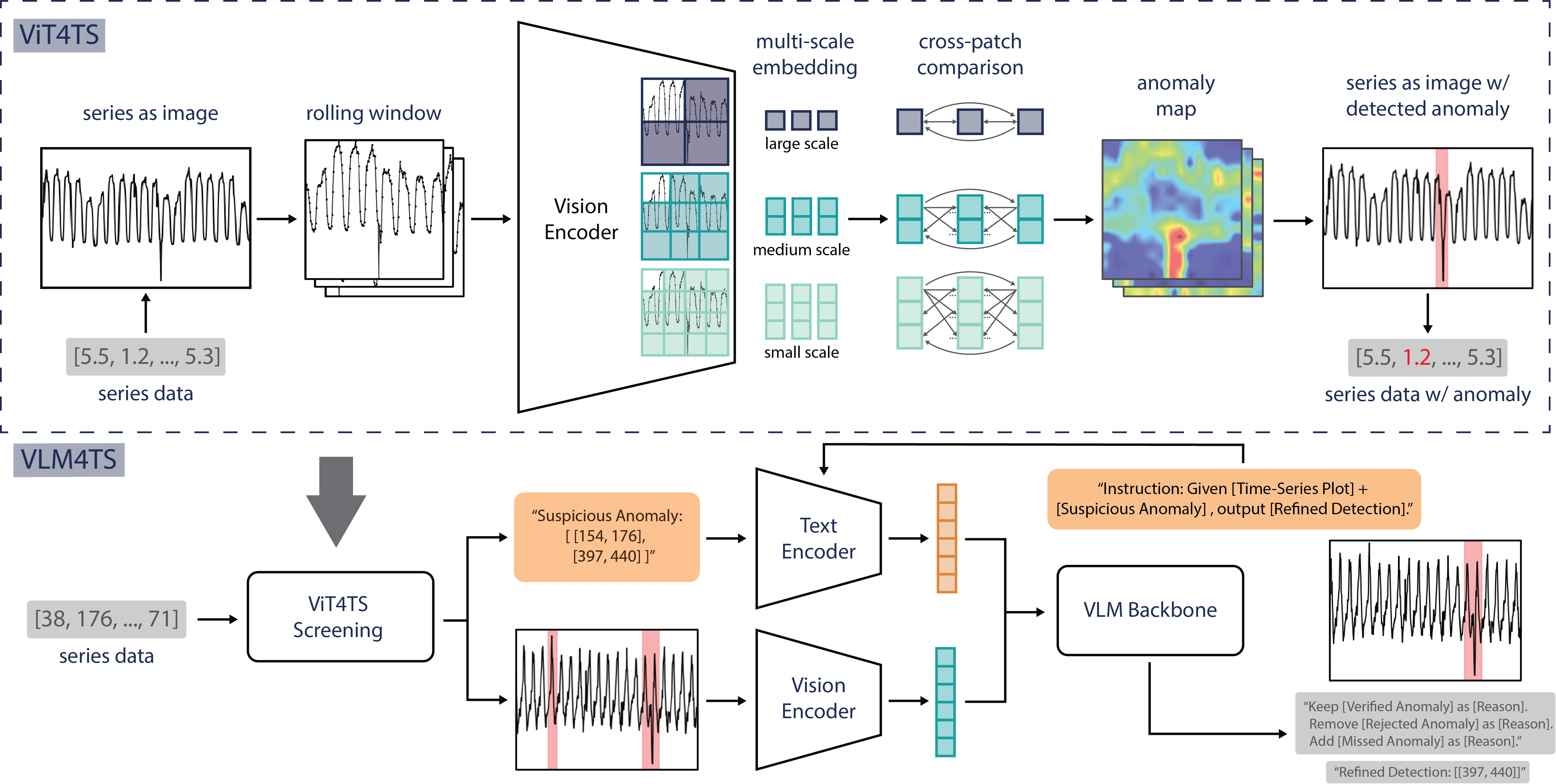}
    \caption{Overview of ViT4TS/VLM4TS (upper/lower pane). In ViT4TS, the raw time series is sliced into windows, and each window is transformed into an image and then embedded into multi-scale feature vectors. By comparing each of these features to others, ViT4TS localizes potentially anomalous regions, and outputs a set of candidate anomaly intervals. Then in VLM4TS, a VLM is then prompted to integrate global temporal context to refine the detection.}
    \label{fig:overview}
\end{figure*}

To overcome the resolution–context dilemma and token inefficiency of naive VLM-based TSAD, we introduce a unified, two-stage framework (Figure \ref{fig:overview}). For the first stage, we propose ViT4TS, which leverages a lightweight, pretrained vision transformer on sliding-window plots to rapidly screen the entire series and generate anomaly candidates via cross-patch comparisons. In the second stage, we propose VLM4TS, which takes each candidate proposal, renders it at a larger temporal scale, and applies an LLM's cross-modal understanding to verify and precisely localize anomalies across extended horizons.

\textbf{Problem Formulation.} 
Consider a univariate time series $\boldsymbol{x}=(x_{1},...,x_{T}) \in \mathbb{R}^T$, where $x_t$ represents the value sampled at timestamp $t$. Assume there exists a set of anomalies of varied length $\mathbf{A}=\{\left(t_s, t_e\right)^i \mid 1 \leq t_s<t_e \leq T\}_{i=1}^m$ that is unknown a priori, our goal is to find a set of $m$ anomalous time intervals, where $t_s$ and $t_e$ represent the start and end time points of an anomalous interval. In this paper, we mainly consider the univariate time series to focus on key method development; a discussion on extending the method for multivariate time series is provided in the Appendix.

\subsection{ViT4TS: Visual Times Series Anomaly Screening}
\label{subsec:vit4ts}

\textbf{Time Series as Line Graphs.} We convert 1-D time series into 2-D images by rendering them as clean line plots, a representation that aligns with both human intuition and the visual pretraining of vision encoders. Line graphs preserve temporal ordering and relative amplitude changes, enabling our ViT4TS module to perform meaningful cross-patch comparisons. As illustrated in Figure \ref{fig:overview}, to minimize non-informative visual artifacts, we omit axis ticks, legends, and other decorative elements. 

\textbf{Rolling Windows and Image Creation.} Pretrained vision encoders usually require square inputs (e.g.\ $224\times224$ or $336\times336$ pixels), yet raw series $\boldsymbol{x}$ often span far more time steps than a single image width. To achieve precise, local-scale anomaly localization, we extract overlapping windows of length $L_w$—chosen to match the image width—using a stride $L_s=\lfloor L_w/4\rfloor$. Each segment is rendered as a clean line plot on an $L_w\times L_w$ canvas, where each time tick maps to one pixel column. 
To make images of different time segments directly comparable, we use the same y-axis limits set to $[\min_t x_t,\max_t x_t]$ for every image. This process produces \(N\approx (T - L_w)/L_s + 1\) images \(\mathbf{I}_i \in \mathbb{R}^{L_w \times L_w \times 3}\) (replicated to three channels for compatibility), which are then fed into vision encoders. For brevity, we omit explicit window subscripts in $\mathbf{I}_i$ in later sections.

\textbf{Multi-Scale Embedding Extraction.} 
For image encoding, we adopt the pretrained CLIP vision encoder \(f_{\mathrm{clip}}\) \cite{schuhmann2021laion}, as it is a standard VLM vision encoder backbone \cite{NEURIPS2023_6dcf277e, Liu_2024_CVPR} and thus can align our visual screening results with the subsequent VLM4TS stage, yet any other vision model can also be used within this framework. To capture fine-grained time series anomalies, following the standard practice in image segmentation \cite{Jeong_2023_CVPR,MishraAD}, for each image we extract the full patch-level penultimate feature map
\(
\mathbf{F} = f_{\mathrm{clip}}(\mathbf{I})\;\in\;\mathbb{R}^{P\times P\times D},
\)
where $P$ is the patch size and $D$ is the embedding dimension. In this way, anomalies like narrow spikes or brief dips can be localized at patch resolution. 

At the same time, many anomalies, such as an extended jump or a drift that distorts the temporal patterns over dozens of time steps, will span multiple patches and demand broader context. To capture features at varying spatial scales, we generate pooled feature maps \(\{\mathbf{F}^{(k)}\}_{k\in\mathcal{K}}\) by applying average pooling with kernel size \(k\) and stride 1 over the base patch feature map \(\mathbf{F}\in\mathbb{R}^{P\times P\times D}\). For each \(k\in\mathcal{K}\) and spatial location \((i,j)\),
\[
\mathbf{F}^{(k)}_{i,j,d}
=
\frac{1}{k^2}
\sum_{u=0}^{k-1}\sum_{v=0}^{k-1}
\mathbf{F}_{i+u,\;j+v,\;d}.
\]
Here \(i,j = 1,\dots,P-k+1\) and \(d=1,\dots,D\).
Each \(\mathbf{F}^{(k)}\in\mathbb{R}^{(P-k+1)\times(P-k+1)\times D}\) aggregates overlapping \(k\times k\) neighborhoods, trading fine spatial detail for broader contextual information.  These scale-specific feature maps are collected for later aggregation. Another advantage of such a multi-scale embedding is to enable rich cross-window comparisons, which in turn allows a larger rolling-window stride $L_{s}$, and thus improving efficiency without sacrificing localization accuracy.

\textbf{Cross-Patch Comparison.}
As we don't have any ground-truth normal reference, we leverage the rarity of anomalies by matching each window’s patch embeddings against those of all other windows at multiple scales. Concretely, for each sliding-window index \(i\) and scale \(k\in\mathcal{K}\), let  
\(\;\mathbf{L}_i^{(k)}\in\mathbb{R}^{P^2\times D}\)  
denote the flattened embedding grid. To score patch \(p\) in window \(i\), we first compute its cosine dissimilarity to every patch \(r\) in each other window \(j\neq i\),  then aggregate across windows via the median, that is, we obtain a cross-patch reference map 
\begin{align*}
&\mathbf{S}_i^{(k)}[p]
=\mathrm{median}_{j\neq i}\,\boldsymbol{U}_{i,j}^{(k)}[p], \ \text{with} \\   &\boldsymbol{U}_{i,j}^{(k)}[p]
=\min_{r}\bigl[1 - \cos(\mathbf{L}_{i}^{(k)}[p],\mathbf{L}_{j}^{(k)}[r])\bigr]. 
\end{align*}
Such a cross-patch matching can flexibly capture local pattern correspondences, regardless of spatial position, making it capable of detecting rare patterns while being robust to irregular seasonal shifts, trends, or periodic motifs. To reduce memory usage, we also evaluate a median-reference variant: we first compute a single reference map  and then score each test patch, that is, 
\begin{align*}
\;
&\tilde{\mathbf{S}}_i^{(k)}[p]
=\min_{r}\bigl[1 - \cos(\mathbf{L}_{i}^{(k)}[p],\mathbf{V}^{(k)}[r])\bigr], \text{with} \\   \ \   &\mathbf{V}^{(k)}[r]=\mathrm{median}_{j}\mathbf{L}_{j}^{(k)}[r].
\end{align*}
The median-reference variant drastically reduces memory usage, while the all-pairs variant remains more sensitive to rare anomalies. In our main experiments, we adopt the median-reference approach for its efficiency and report the min-reference results in the Appendix. Finally, to produce a single, fine-grained anomaly map, we upsample each scale-specific map \(\{\mathbf{S}_i^{(k)}\}_{k\in\mathcal{K}}\) to the base patch resolution and fuse them via harmonic averaging at every patch location. This multi-scale fusion combines the pinpoint sensitivity of small-scale scores with the broader context captured at larger scales. Examples of the resulting aggregated patch-level anomaly maps are shown in Figure \ref{fig:overview}. 

\textbf{Anomaly Scoring and Localization.} 
To assemble a final anomaly score per time step, we first map each fused patch-level score back to its original time index and average overlapping contributions to form a 2-D anomaly map \(\mathbf{M}\in\mathbb{R}^{P\times T}\), where \(P\) is the patch resolution and \(T\) the series length. We then collapse \(\mathbf{M}\) to a 1-D score \(s\in\mathbb{R}^T\) by taking the \(q\)-th quantile across all rows:  
\(
s(t)=\mathrm{quantile}_{q}\bigl(\mathbf{M}_{:,t}\bigr).
\)  
Lower \(q\) (e.g.\ 0.1) increases sensitivity to subtle distortions, while higher \(q\) (e.g.\ 0.5) emphasizes large spikes; we set \(q=0.25\) for a balanced trade-off (See the Appendix for details). Finally, we choose a threshold \(\tau\) at the \((1-\alpha)\)-Gaussian quantile of \(s\) and extract all contiguous intervals \(\hat{\mathbf{A}}=\{(t_s,t_e)^i\mid s(t)>\tau,\ \forall\,t_s\le t\le t_e\}_{i=1}^m\) as our detected anomalies from ViT4TS.  

\subsection{VLM4TS: Anomaly Verification and Refinement}
\label{subsec:vlm4ts}

Once ViT4TS produces a set of accurate candidate intervals, we send the full series image and these candidate proposals to VLM4TS for verification under the global context. ViT4TS excels at precise, local detection but may flag benign fluctuations or miss extended anomalies; VLM4TS uses cross-modal understanding along with the global temporal context to resolve these cases. In our main experiments, we employ GPT-4o, and evaluate alternatives in ablation studies.

\textbf{Visual Input.} We render each complete time series as an ordinary line graph—with evenly spaced x-axis ticks and y-axis value labels—so that the VLM can perceive trends, seasonality, and drifts at a glance. Because ViT4TS has already provided precise endpoints, we feed this single, full-length image per signal into the VLM’s vision channel, avoiding further windowing and enabling understanding over the entire horizon.

\textbf{Textual Input.} In parallel, we supply a prompt that (1) lists the initial proposals $\hat{\mathbf{A}}$ and remind the model these were generated from local shape matching; (2) instructs it to confirm only intervals that truly deviate from the global pattern; (3) rejects any false positives consistent with overall behavior; and (4) suggests additional intervals exhibiting clear statistical or visual irregularities that ViT4TS may have missed. We further ask the VLM to assign each interval a confidence score from 1 (low) to 3 (high). See the Appendix for the full prompt. The VLM returns a JSON object containing the refined anomaly set \(\hat{\mathbf{A}}_{final}=\{(t_s,t_e)^i\}_{i=1}^{\hat m}\), per-interval confidence ratings, and a brief natural-language justification for each decision. We discard any interval with confidence = 1 to produce our final detection and diagnosis.

\textbf{Remark on Computation.} By confining expensive VLM inference in a longer context rather than every sliding window, our two-stage pipeline cuts token usage by an average factor of 36 $\times$ compared to naive rolling-window prompting (Table \ref{tab:F1versusLLM}). ViT4TS (Stage 1) runs in reasonably short time with just CPU for moderate-length series (See the Appendix), and VLM4TS (Stage 2) incurs only a few seconds of latency per series—well within acceptable bounds for offline analysis or human-in-the-loop systems. 

\section{Experiments}
\label{sec:exp}
We perform an array of experiments to evaluate the performance of ViT4TS and VLM4TS on a variety of benchmarks on industrial anomaly detection. We also conduct an extensive ablation study to validate the individual effectiveness of our proposed components. Detailed experimental setups—including data preprocessing, evaluation metrics are provided in the Appendix. 

\subsection{Experimental Setup}

\textbf{Benchmark Datasets and Baseline Methods.} For unsupervised anomaly detection, we follow the standard evaluation protocol to test ViT4TS and VLM4TS on 11 widely used benchmark datasets in time series anomaly detection research \cite{lavin2015evaluating,hundman2018lstmdt}, spanning various domains from sensor data like astronomy sensory, web monitoring data like production traffic, to web metric data like volume of Twitter mentions, to evaluate the models’ generalizability and adaptability. For baseline models, we compare our method against several anomaly detection approaches, from \textit{statistical baselines} like \texttt{ARIMA} \cite{pena2013arima}, to deep learning models currently considered state-of-the-art, including a forecasting- based LSTM model (\texttt{LSTM-DT}) \cite{hundman2018lstmdt}, (variational) reconstruction-based models like \texttt{LSTM-AE} \cite{malhotra2016lstmae}, \texttt{VAE} \cite{park2018vae}, and \texttt{TadGAN} \cite{geiger2020tadgan}, hybrid models like \texttt{AER} \cite{wong2022aer}, transformer-based models like Anomaly Transformer (\texttt{ATrans}) \cite{xu2021anomaly}, and pre-trained time series foundation models like \texttt{UniTS} \cite{gao2024units} and \texttt{TimesFM} \cite{das2024decoder}, as well as LLM-based approaches such as prompt-based detectors (\texttt{SigLLM‐P\(_{\mathrm{G}}\)} on GPT, \texttt{SigLLM‐P\(_{\mathrm{M}}\)} on Mistral) and LLM prediction-based models \texttt{SIGLLM-D} \cite{alnegheimish2024can}. We also report \texttt{TAMA} \cite{zhuang2024see}, a naive rolling-window VLM prompting baseline.

\textbf{Evaluation.}
For any method that produces continuous anomaly scores, we first smooth the raw outputs with an exponentially weighted moving average. We then apply a Gaussian-based threshold of the form \(\mu + k\sigma\), where \(\mu\) and \(\sigma\) are the mean and standard deviation of the smoothed scores, sweeping \(k\) to compute the unweighted contextual F1 score \cite{geiger2020tadgan,wong2022aer}.
Max F1 score (F1-max) results appear in the main text; full F1 results are provided in the Appendix. Methods yielding only binary labels are evaluated at their default settings and reported by raw F1. See Appendix for the formal definition of these metrics. All experiments run on an NVIDIA V100 GPU if not otherwise specified. Baselines are implemented via the Orion framework \cite{alnegheimish2022orion} when available; otherwise, we follow each original implementation’s setup and prompt.  

vs. detection delay).
\begin{table*}[ht]
  \captionsetup{skip=4pt}  
  \caption{Detection performance (F1-max) of ViT4TS and VLM4TS versus trained-from-scratch and time-series-pretrained baselines on benchmark datasets. Each entry reports the maximum F1 score across all evaluated thresholds; the best score is shown in bold, and the second-best is underlined. Definition of the F1-max score, full F1 results and elapsed (wall-clock) time comparison are provided in the Appendix.}
\centering
\label{tab:F1versusTS}
\resizebox{\linewidth}{!}{%
\begin{tabular}{llcccccccccccc}
\toprule
\textbf{Type}& \textbf{Method}& \multicolumn{5}{c}{\textbf{NAB}} & \multicolumn{2}{c}{\textbf{NASA}} & \multicolumn{4}{c}{\textbf{YAHOO}} & \\
      \cmidrule(lr){3-7}\cmidrule(lr){8-9}\cmidrule(lr){10-13} 
                  &  & \textbf{Art} & \textbf{AWS} & \textbf{AdEx} & \textbf{Traf} & \textbf{Tweets} & \textbf{MSL} & \textbf{SMAP} & \textbf{A1} & \textbf{A2} & \textbf{A3} & \textbf{A4} & \(\boldsymbol{\mu\pm\sigma}\) \\
\midrule
\multirow{7}{*}{\makecell[l]{Trained\\From\\Scratch}}
& \texttt{ARIMA}      & 0.387 & 0.263 & 0.500 & 0.344 & 0.179 & \underline{0.585} & 0.750 & \underline{0.650} & 0.771 & 0.502 & 0.336 & 0.479$\pm$0.187 \\
& \texttt{AER}        & 0.338 & 0.244 & 0.518 & 0.404 & 0.178 & 0.553 & \underline{0.753} & 0.618 & 0.866 & \textbf{0.711} & \textbf{0.614} & 0.527$\pm$0.207 \\
& \texttt{TadGAN}     & 0.338 & 0.196 & 0.385 & 0.421 & 0.205 & 0.612 & 0.593 & 0.492 & 0.667 & 0.135 & 0.109 & 0.378$\pm$0.189 \\
& \texttt{LSTM-DT}    & 0.368 & 0.273 & 0.444 & 0.451 & 0.190 & 0.615 & 0.724 & 0.639 & 0.877 & \underline{0.704} & 0.534 & 0.529$\pm$0.200 \\
& \texttt{LSTM-AE}    & 0.231 & 0.244 & 0.400 & 0.416 & 0.232 & 0.487 & 0.673 & 0.583 & 0.853 & 0.584 & 0.201 & 0.446$\pm$0.203 \\
& \texttt{ATrans}     & 0.262 & 0.168 & 0.200 & 0.365 & 0.147 & 0.454 & 0.567 & 0.263 & 0.554 & 0.437 & 0.394 & 0.346$\pm$0.142 \\
& \texttt{VAE}        & 0.000 & 0.248 & 0.345 & 0.323 & 0.237 & 0.515 & 0.695 & 0.557 & 0.845 & 0.524 & 0.189 & 0.407$\pm$0.234 \\
\midrule
\multirow{3}{*}{\makecell[l]{Time-series\\Pretrained}}
& \texttt{UniTS}      & 0.182 & 0.246 & 0.326 & 0.479 & 0.167 & 0.561 & 0.723 & 0.605 & 0.760 & 0.126 & 0.110 & 0.390$\pm$0.233 \\
& \texttt{TimesFM}    & 0.234 & 0.243 & 0.400 & 0.467 & 0.198 & 0.564 & 0.686 & 0.554 & 0.694 & 0.120 & 0.107 & 0.388$\pm$0.209 \\
& \texttt{TimesFM2} 
& 0.255 & 0.233 & 0.364 & 0.386 & 0.171 
& 0.556 & 0.676 
& 0.594 & 0.692 & 0.202 & 0.181 
& 0.392$\pm$0.194 \\
\midrule
\multirow{2}{*}{Ours}
& \textbf{ViT4TS}  & \underline{0.545} & \underline{0.400} & \underline{0.615} & \underline{0.615} & \underline{0.597} & 0.543 & 0.726 & 0.614 & \underline{0.892} & 0.614 & \underline{0.565} & \underline{0.612$\pm$0.116} \\
& \textbf{VLM4TS}  & \textbf{0.714} & \textbf{0.488} & \textbf{0.727} & \textbf{0.632} & \textbf{0.686} & \textbf{0.619} & \textbf{0.773} & \textbf{0.733} & \textbf{0.901} & 0.497 & 0.474 & \textbf{0.659$\pm$0.127} \\
\bottomrule
\end{tabular}%
}
\end{table*}

\begin{table*}[t]
\centering
  \captionsetup{skip=4pt}
\caption{Performance and efficiency comparison of VLM4TS versus language model–based baseline methods on benchmark datasets. “Tokens” reports the average number of tokens consumed per time series; “Time” indicates the average elapsed (wall-clock) time to generate detections per time series. Efficiency metrics have been adjusted to account for the methods' window and step size difference.}
\label{tab:F1versusLLM}
\resizebox{\linewidth}{!}{%
\begin{tabular}{llccc ccc ccc ccc}
\toprule
\textbf{Type} & \textbf{Method}
  & \multicolumn{3}{c}{\textbf{NAB}}
  & \multicolumn{3}{c}{\textbf{NASA}}
  & \multicolumn{3}{c}{\textbf{YAHOO}}
  & \multicolumn{3}{c}{$\boldsymbol{\mu}$} \\
\cmidrule(lr){3-5}\cmidrule(lr){6-8}\cmidrule(lr){9-11}\cmidrule(lr){12-14}
 & & F1‐max & Tokens & Time
 & F1‐max & Tokens & Time
 & F1‐max & Tokens & Time
 & F1‐max & Tokens & Time \\
\midrule
\multirow{3}{*}{LLM‐based}
 & \texttt{SIGLLM‐D}        
   & 0.353 & \underline{11153} & \underline{83.77} 
   & 0.232 & \underline{13157} & 89.50 
   & 0.393 & \underline{21508} & 62.20 
   & 0.326 & \underline{15273} & \underline{78.49} \\
 & \texttt{SigLLM‐P}\(_{\mathrm{M}}\) 
   & 0.206 & 14191 & 613.28 
   & 0.157 & 45050 & 2107.96 
   & 0.276 & 45082 & 984.45 
   & 0.213 & 34774 & 1235.23 \\
 & \texttt{SigLLM‐P}\(_{\mathrm{G}}\) 
   & 0.162 & 25006 & 2258.78 
   & 0.080 & 78187 & 2614.87 
   & 0.143 & 83207 & 2852.52 
   & 0.128 & 62133 & 2575.39 \\
\midrule
\multirow{1}{*}{VLM‐based}
 & \texttt{TAMA}           
   & \underline{0.513} & 40009 & 110.47 
   & \underline{0.631} & 31563 & \underline{83.49} 
   & \underline{0.616} & 27324 & 68.99 
   & \underline{0.587} & 32965 & 87.65 \\
\midrule
\multirow{1}{*}{Ours}
 & \textbf{VLM4TS} 
   & \textbf{0.649} & \textbf{1219} & \textbf{16.71}
   & \textbf{0.696} & \textbf{1213} & \textbf{20.97}
   & \textbf{0.651} & \textbf{1204} & \textbf{6.36}
   & \textbf{0.665} & \textbf{1212} & \textbf{14.68} \\
\bottomrule
\end{tabular}%
}
\end{table*}


\subsection{Performance Evaluation}

\textbf{Overall Detection Performance.}  
Table \ref{tab:F1versusTS} reports the F1-max scores versus trained-from-scratch and time-series-pretrained baselines on all 11 benchmark datasets. Our two-stage VLM4TS framework achieves the highest average F1-max, outperforming competing methods on 9 out of 11 datasets, achieving a 24.6 \% improvement in average F1-max score over the second-best baseline \texttt{LSTM-DT}. These results underscore that purely vision-driven screening, when coupled with powerful VLM understanding, can exceed state-of-the-art time series-based TSAD models in most cases. Furthermore, even the first-stage visual screening module ViT4TS ranks second overall, securing the top-two position on 6 datasets, showing the effectiveness of detecting time series anomalies from a visual perspective. In Table \ref{tab:F1versusLLM}, we further compare against existing language model-based baseline methods, where VLM4TS also consistently outperform baselines by a large margin, achieving 13.3\% performance gain over the VLM prompting-based method \texttt{TAMA} and $\times2$ improvement over the strongest LLM-based method \texttt{SigLLM-D}. Moreover, we observe that both VLM-based methods substantially surpass LLM prompting frameworks, confirming that casting time series as images unlocks more effective anomaly detection than text-only representations.

\textbf{Task-Specific Detection Performance.} Compared with time-series-based methods in Table~\ref{tab:F1versusTS}, VLM4TS delivers its strongest gains on real-world datasets dominated by contextual anomalies (e.g., the NAB dataset group), where it outperforms all baselines by a large margin. Its advantage stems from the ability to localize anomalies in a 2-D representation and then apply global-context verification to boost precision without sacrificing recall {(see discussion on precision and recall in the Appendix)}.
In contrast, on a few synthetic datasets, such as A3 and A4, where anomalies are densely and synthetically injected (average anomalies per time series: A3 = 9.39, A4 = 8.37 versus A1 = 2.66, A2 = 2.00), we see forecasting-based methods (e.g., \texttt{AER} and \texttt{LSTM-DT}) excelling compare to our method. {Because VLM4TS assumes anomalies are rare, it adopts a more conservative behavior in high-density scenarios: visually similar, tightly packed anomaly points are interpreted as fluctuations within a broader pattern rather than as anomalies. This conservativeness results in fewer selected anomalies, leading to lower F1 scores on A3 and A4 (see Appendix for further discussion).} Compared with language-based methods shown in Table~\ref{tab:F1versusLLM}, such as \texttt{TAMA}, VLM4TS delivers the strongest gains in datasets that demand long-range context, most notably the NAB dataset group (26.5\%). This advantage arises from our two-stage design: ViT4TS supplies high-resolution, localized proposals, allowing VLM4TS to verify and refine anomalies over a much larger temporal horizon without losing localization accuracy.

\textbf{Computational Time and Token Cost.}
Table \ref{tab:F1versusLLM} shows the F1-max score, per-series elapsed time, and token consumption for VLM4TS and language model-based baseline methods. We also report per-series elapsed time comparison with time-series-based method in the Appendix. Due to our two-stage design, VLM4TS requires substantially less cost, reducing token usage by an average of 30× compared to pure LLM- and VLM-based detectors that encode numerical data into text or image for every rolling window. When compared against time series–only approaches—both pretrained foundation models and models trained from scratch, VLM4TS achieves comparable end-to-end runtime while achieving better overall detection accuracy. We also evaluate CPU-only operation of VLM4TS (with VLM running on the API) with the default ViT-B/16 backbone. On moderate-length series spanning thousands of time points, ViT4TS screening completes in seconds, demonstrating feasibility for large-scale deployment without GPU acceleration and with low token cost.

\subsection{Additional Analysis}
\textbf{Ablation Study.} Table \ref{tab:ablation_visionad} evaluates the contribution of each ViT4TS component and the necessity of visual screening for VLM4TS. We have the following observations:\\
$\bullet$ \textit{Patch-level embedding:} replacing the detailed grid of patch embeddings with a single global summary embedding ([CLS] token) per window leads to a substantial drop by 11.94\% on average on the benchmark datasets. This confirms that fine-grained, patch-level representations are critical for accurately localizing distortions in time series plots. \\
$\bullet$ \textit{Cross-patch matching}: compared to position-aligned patches comparison (No cross-patch comparison) and row-wise patches comparison (No column-wise comparison), cross-patch matching leads to an substantial 18.76\% and 30.54\% improvement on YAHOO dataset group. This is because a flexible matching across both spatial dimensions is especially important on YAHOO dataset group, where seasonal and trend patterns can hide anomaly patterns. \\
$\bullet$ \textit{Multi-scale embedding:} ablating multi-scale embedding extraction reduces performance, especially on NASA dataset group, which can lead to a 8.35\% drop in F1-max, as multi-scale embeddings enhance detection to extended contextual anomalies that cannot be captured in a patch-level, particularly in domains like spacecraft telemetry. \\
$\bullet$ \textit{Visual screening:} omitting the ViT4TS proposal stage and applying VLM4TS directly to full-series images causes a dramatic F1 collapse, 
especially on dataset with more dense anomalies like those in the YAHOO dataset group. This shows the importance of high-recall, local screening step for VLM to achieve superior performance; without it, the model cannot reliably isolate multiple anomalies within complex temporal backgrounds. 

\begin{table}
  \captionsetup{skip=4pt}
  \caption{Ablation study of ViT4TS and VLM4TS on different dataset groups, reporting F1-max scores for each configuration.}
  \label{tab:ablation_visionad}
  \centering
  \resizebox{\linewidth}{!}{%
    \begin{tabular}{lccc}
      \toprule
      \textbf{Method}               & \textbf{NAB} & \textbf{NASA} & \textbf{YAHOO} \\
      \midrule
      w/o patch-level embedding          & 0.519        & 0.578         & 0.541          \\
      w/o cross-patch comparison    & 0.504        & 0.613         & 0.514          \\
      w/o column-wise comparison      & 0.523        & 0.624         & 0.565          \\
      w/o multi-scale embedding     & 0.534        & 0.582         & \textbf{0.677} \\
      \textbf{ViT4TS (ours)}        & \textbf{0.555}& \textbf{0.635}& \underline{0.671}\\
      \midrule
      w/o ViT4TS                    & 0.539        & 0.517         & 0.292          \\
      \textbf{VLM4TS (ours)}        & \textbf{0.649}& \textbf{0.696}& \textbf{0.651}\\
      \bottomrule
    \end{tabular}%
  }
\end{table}

\begin{figure}[htbp]
    \centering
    \includegraphics[width=\linewidth]{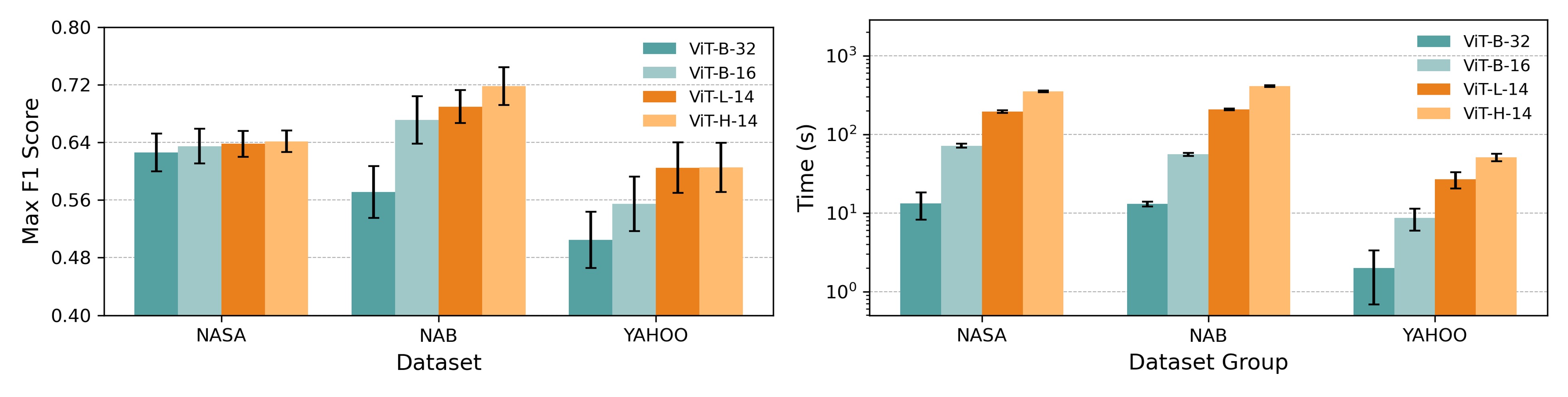}
    \caption{F1-max score and elapsed (wall-clock) time for ViT4TS with different backbones. Error bars indicate standard deviation across 3 replications.}
    \label{fig:vit_backbone}
\end{figure}

\textbf{Backbone Scalability and Efficiency.}
To evaluate the impact of vision encoder capacity and patch resolution on screening performance, we replace our default ViT-B/16 backbone with other backbones within the ViT4TS pipeline and measure both F1-max and inference time (Figure \ref{fig:vit_backbone}). To accommodate real-world industrial environments where GPU resources may be limited, here we evaluate ViT4TS's elapsed time on CPU (Xeon E5, 32 GB RAM). Compared to the default backbone, coarsening the patch grid to 32×32 in ViT-B/32 yields much lower F1-max on NAB (–14.89\%) and YAHOO (–8.99\%), showing the importance of fine patch-level representations for accurate localization. Conversely, increasing model depth and hidden dimension improves detection on volatile series such as those in the YAHOO dataset group, showing that larger backbones better capture complex temporal patterns. However, these gains come at a steep computational cost: For the YAHOO dataset group, ViT-B/32 completes screening in under 2s per series on CPU, whereas ViT-H/14 exceeds 50s on average. Overall, the default backbone ViT-B/16 provides a nice trade-off between accuracy and efficiency for TSAD tasks.  

\begin{figure}
    \centering
    \includegraphics[width=\linewidth]{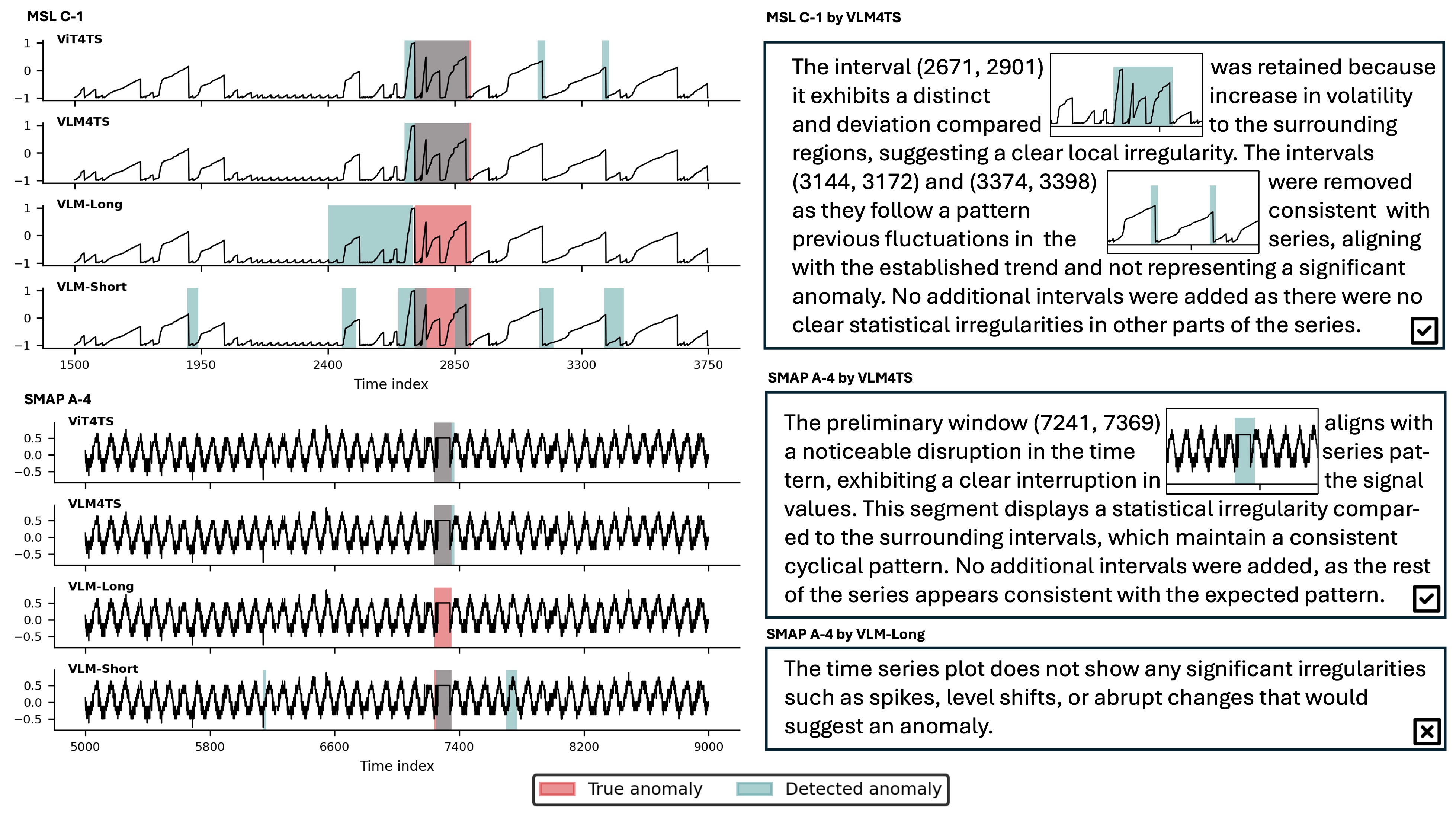}
    \caption{Qualitative results on MSL C-1 (top) and SMAP A-4 (bottom), illustrating how VLM4TS refines the initial local anomaly proposals from ViT4TS by selectively keeping, discarding, or adding intervals based on global temporal understanding. "VLM-Long” refers to VLM prompting on the full-series image (ablation without ViT4TS screening); “VLM-Short” (\texttt{TAMA}) refers to VLM prompting on the rolling-windows image. Only representative segments are shown for clarity.}
    \label{fig:case_study}
\end{figure}

\textbf{Qualitative Results.} Figure \ref{fig:case_study} illustrates anomaly localization on two NASA telemetry signals. In both cases, VLM4TS delivers the most precise detections: it refines ViT4TS’s candidate proposals by comparing each candidate against the full-series context, correctly isolating the true anomalous intervals. By contrast, the “VLM-Long” ablation (prompting the VLM on the entire series without prior screening) either misaligns its anomaly boundaries (MSL C-1) or fails to flag the event entirely (SMAP A-4). Meanwhile, “VLM-Short” (\texttt{TAMA}), which prompts on every rolling window, generates numerous false positives due to its narrow context and incurs prohibitively high token usage.

\textbf{Additional Analyses.} In the Appendix, we provide further analysis of the framework, including patch‐level embedding and anomaly map visualizations, comparisons across different VLM backbones, alternative visualizations such as spectrogram baselines, and a study of how window size and patch scale affect fine‐grained detection, among other results and discussions.

\section{Conclusion}
 We introduce a novel two‐stage framework for unsupervised  time‐series anomaly detection that reframes 1-D time series as 2-D visual inputs and leverages pretrained vision and vision–language models without any in‐domain training. In the first stage, ViT4TS applies multi‐scale cross‐patch comparisons to generate anomaly candidates in high resolution, and in the second stage, VLM4TS verifies these candidate proposals via temporal understanding with long context. Our method achieves state‐of‐the‐art performance on diverse benchmarks while using many fewer tokens.

\section{Acknowledgments}
{We thank Dr.~Runze Li for valuable discussions and suggestions that contributed to the methodological formulation.}

\bibliography{myrefs}

\newpage
\appendix
\onecolumn       

\section{Experimental Details and Further Discussion}
\label{app:exp_details}

\subsection{Benchmark Datasets}
\label{app:benchmark}
We use 11 datasets spanning various domains to evaluate the models' generalizability and adaptability. The National Aeronautics and Space Administration (NASA) provided two spacecraft telemetry datasets \footnote{NASA data: https://github.com/khundman/telemanom/}: Soil Moisture Active Passage (SMAP) and Mars Science Laboratory (MSL) acquired from a satellite and a rover, respectively \cite{hundman2018lstmdt}. Each numeric measurement in the target channel is accompanied by one-hot encoded information about commands sent or received by specific spacecraft modules in a given time window. The Yahoo Webscope Program provided the S5 datasets \footnote{YAHOO data: https://webscope.sandbox.yahoo.com/catalog.php?datatype=s\&did=70} consisting of one set of real production traffic to Yahoo properties (A1) and three synthetic datasets (A2, A3, A4) with varying trends, noise, and pre-specified or random seasonality. The A2 and A3 datasets only contain outliers inserted at random positions, while A4 has outliers and change points. The Numenta Anomaly Benchmark (NAB) provided several datasets \footnote{NAB data: https://github.com/numenta/NAB} from various domains: artificialWithAnomaly (Art), realAdExchange (AdEx), realAWSCloudwatch (AWS), realTraffic (Traffic), realTweets (Tweets). NAB is challenging due to its variety and the requirement for detectors to work across domains. Results on NAB can assess how well our approach handles completely different kinds of time series in one unified model. Similar to \cite{wong2022aer}, Table \ref{tab:dataset_properties} summarizes the basic information of each dataset. It differentiates between real and synthetic datasets and provides the number of
anomalies for each dataset. Each anomaly is classified as either
point or collective, depending on the length of the anomaly.
Lastly, the total number of anomalous and overall data points
are provided for each dataset.

\begin{table*}[ht]
\centering
\caption{Summary of dataset properties across benchmarks.}
\label{tab:dataset_properties}
\resizebox{\linewidth}{!}{%
\begin{tabular}{ll cc cccc ccccc}
\toprule
\multirow{2}{*}{Category} & \multirow{2}{*}{Metric}
  & \multicolumn{2}{c}{NASA}
  & \multicolumn{4}{c}{YAHOO}
  & \multicolumn{5}{c}{NAB} \\
\cmidrule(lr){3-4} \cmidrule(lr){5-8} \cmidrule(lr){9-13}
  & & MSL & SMAP & A1 & A2 & A3 & A4 & Art & AdEx & AWS & Traf & Tweets \\
\midrule
Properties   & Synthetic \# signals
  & No & No & No & Yes & Yes & Yes & Yes & No & No & No & No \\
\addlinespace
\# Anomalies  & Point (len = 1)
  & 0 & 0 & 68 & 33 & 935 & 833 & 0 & 0 & 0 & 0 & 0 \\
             & Collective (len $>$ 1)
  & 36 & 67 & 110 & 167 & 4 & 2 & 6 & 11 & 30 & 14 & 33 \\
\addlinespace
\# Data Points & Anomalous points
  & 7766 & 54696 & 1669 & 466 & 943 & 837 & 2418 & 795 & 6312 & 1560 & 15651 \\
               & Total points
  & 132046 & 562800 & 94866 & 142100 & 168000 & 168000 & 24192 & 7965 & 67644 & 15662 & 158511 \\
\bottomrule
\end{tabular}%
}
\end{table*}

\subsection{Model Setup} For the primary experiment, ViT4TS uses the OpenCLIP \cite{ilharco_gabriel_2021_5143773} implementation of CLIP with the ViT-B/16 backbone, pretrained on the LAION-400M dataset \cite{schuhmann2021laion}. For multi-scale embedding extraction, we perform average pooling over the patch grid using kernel sizes of \(2\times2\) and \(3\times3\). For the verification stage, VLM4TS employs GPT-4o (API version 2024-08-06) via OpenAI’s API.  

\subsection{Baseline Methods and Evaluation Procedure}
\label{app:baseline}
We evaluate our methods against a broad spectrum of TSAD baselines, as detailed below:

\noindent\textbullet\ \texttt{ARIMA} \citep{pena2013arima}: autoregressive integrated moving average forecasting model; learns autocorrelations to predict future values and uses point‐wise forecast errors as anomaly scores.\\
\noindent\textbullet\ \texttt{LSTM‐DT} \citep{hundman2018lstmdt}: two‐layer LSTM for one‐step ahead prediction; uses residuals as anomaly scores.\\
\noindent\textbullet\ \texttt{LSTM‐AE} \citep{malhotra2016lstmae}: LSTM autoencoder over sliding windows; use reconstruction error as anomaly scores.\\
\noindent\textbullet\ \texttt{VAE} \citep{park2018vae}: variational autoencoder with LSTM encoder/decoder; use  reconstruction likelihood and point‐wise error to construct anomaly scores.\\
\noindent\textbullet\ \texttt{AER} \citep{wong2022aer}: joint autoencoder and bidirectional LSTM predictor; optimizes reconstruction and forward/backward prediction, combining DTW‐based and point‐wise errors for anomaly scores.\\
\noindent\textbullet\ \texttt{TadGAN} \citep{geiger2020tadgan}: GAN with LSTM generator and critic; computes anomaly score via dynamic‐time‐warping, point‐wise, and area‐difference errors between real and generated series.\\
\noindent\textbullet\ \texttt{ATrans} ({Anomaly Transformer}) \citep{xu2021anomaly}: transformer with anomaly‐attention mechanism measuring association discrepancies among adjacent points as anomaly scores.\\
\noindent\textbullet\ \texttt{UniTS} \citep{gao2024units}: unified multi‐task transformer pretrained on diverse series; tokenizes forecasting, imputation, and anomaly tasks for cross-domain transfer. Using forecasting error as anomaly scores.\\
\noindent\textbullet\ \texttt{TimesFM} / \texttt{TimesFM2} \citep{das2024decoder}: decoder‐style attention model pretrained on patched time-series corpus; using forecasting error as anomaly scores.\\
\noindent\textbullet\ \texttt{SigLLM‐P} (\texttt{SigLLM‐P}\(_{\mathrm{G}}\)  on GPT, \texttt{SigLLM‐P}\(_{\mathrm{M}}\)  on Mistral) \citep{alnegheimish2024can}: prompt‐based detectors that ask an LLM to mark anomalous indices in textual form.\\
\noindent\textbullet\ \texttt{SigLLM‐D} \citep{alnegheimish2024can}: detects anomalies via LLM forecasting errors—comparing next‐step predictions to actual values.\\
\noindent\textbullet\ \texttt{TAMA} \citep{zhuang2024see}: rolling‐window VLM prompts, rendering each window as an image and querying GPT-4o for anomaly detection, using confidence score as anomaly scores.\\

All baselines except \texttt{SigLLM} and \texttt{TAMA} are evaluated using the Orion pipeline \footnote{Orion: https://github.com/sintel-dev/Orion}.  \texttt{SigLLM} is run via its repository \footnote{SIGLLM: https://github.com/sintel-dev/sigllm}, while \texttt{TAMA}—which lacks a public implementation—is reproduced using the original prompts. \texttt{TAMA} is implemented on 20\% of the series,  due to high API costs.

\subsection{Data Preprocessing, Postprocessing and Evaluation.} 
Each raw series \(x_{1:T}\) is first min–max normalized to \([0,1]\) and detrended by removing a least‐squares linear fit; for the A4 dataset, we additionally standardize before and after the known change point. After generating anomaly scores, we apply an exponentially weighted moving average (EWMA) to smooth each score \cite{alnegheimish2022sintel,alnegheimish2022orion, hundman2018lstmdt}. We then apply Gaussian quantile thresholds
\[
\tau_\alpha \;=\; \mu + k\sigma \;=\; \mu + z_\alpha\,\sigma,
\]
where \(\mu\) and \(\sigma\) are the mean and standard deviation and \(z_\alpha\) is the standard normal deviate at quantile \(\alpha\in\{0.10,0.01,0.001\}\), and report the unweighted contextual F1 and the maximum F1 (\(\mathrm{F1\text{-}max}\)) over these \(z_{\alpha}\) values \citep{hundman2018lstmdt,wong2022aer,geiger2020tadgan}. Formally, let the ground-truth anomaly intervals be
\[
\mathbf{A} = \{\bigl(t_s,\;t_e\bigr)^i\}_{i=1}^{m},
\quad
\hat{\mathbf{A}} = \{\bigl(\hat t_s,\;\hat t_e\bigr)^j\}_{j=1}^{n}.
\]
Define
\[
\mathrm{TP}
= \left|\bigl\{\,j : \exists\,i,\;\bigl(\hat t_s,\hat t_e\bigr)^j \cap \bigl(t_s,t_e\bigr)^i \neq \emptyset\bigl\}\right|,
\]
\[
\mathrm{FP}
= \left|\bigl\{\,j : \forall\,i,\;\bigl(\hat t_s,\hat t_e\bigr)^j \cap \bigl(t_s,t_e\bigr)^i = \emptyset\bigl\}\right|,
\]
\[
\mathrm{FN}
= \left|\bigl\{\,i : \forall\,j,\;\bigl(t_s,t_e\bigr)^i \cap \bigl(\hat t_s,\hat t_e\bigr)^j = \emptyset\bigl\}\right|.
\]
The unweighted contextual precision, recall, and F1 score are then calculated corespondingly. No interval padding is applied, ensuring that the detection performance reflects each method’s raw localization capability.

\subsection{Prompt Design for VLM4TS}
\label{app:prompt_design}
We prompt the VLM with the full-series line plot and indicate the intervals  ViT4TS marked as anomalous. Each candidate anomaly is precisely localized through the x-axis tick marks. Our goal for the VLM is to refine the output of the ViT4TS and increase the precision scores. It directs the model to discard any intervals that align with the overall trend, add missed segments exhibiting clear spikes or level shifts, and retain the exact tick-based boundaries from ViT4TS. A three-level confidence scale (1–3) supports downstream filtering. See the attached prompting box for details.

\subsection{Limitation and Future Work.} While our results demonstrate VLMs’ promise for unsupervised anomaly detection, our current prompting strategy remains simple. Future work could explore advanced in-context reasoning methods, such as chain-of-thought or retrieval-augmented prompts, to further enhance temporal inference. Additionally, we have focused on univariate series; extending to multivariate data is an interesting future direction. In the following section, we give an initial exploration of a multivariate extension, showcasing how the two-stage design can be extended to incorporate multi-channel temporal information.

\section{Additional Experimental Results}
\label{app:exp_results}

\subsection{Full Ablation Analysis}
\label{app:full_ablation}

Table \ref{tab:ablation_full_scores} presents the full ablation results, extending the analyses from Section 4.3 with two additional experiments. First, we evaluate an \emph{all‐pairs} cross‐patch comparison—computing the minimum cosine dissimilarity between every patch in the test window and every patch across all other windows (instead of using a single median reference). As expected, all‐pairs matching yields higher F1‐max across multiple threshold quantiles, though at the expense of an increased memory usage (as we need to store embeddings for all windows). The default median‐reference approach (reported in the main paper) therefore remains preferable for its efficiency. 

Second, we vary the quantile \(q\) used to collapse the 2-D anomaly map to 1-D (Section 3.1). In addition to the standard \(q=25\%\), we test \(q=12.5\%\) and \(q=50\%\). Reducing \(q\) to 12.5\%\ produces nearly identical F1‐max, while increasing \(q\) to 50\%\ improves detection on signals with extended, high‐amplitude anomalies (e.g.\ long spikes). All other variants—patch‐level embedding, cross‐patch matching ablations, multi‐scale embedding, and omission of the ViT4TS screening stage—reproduce the trends reported in Section 4.3, confirming that (i) fine‐grained patch features, (ii) flexible cross-patch comparision, (iii) multi‐scale feature extraction  are each crucial to our performance gains.  

\begin{table}[ht]
  \captionsetup{skip=4pt}
  \caption{Full F1 scores of ViT4TS ablation variants at different thresholding quantile $\alpha$, grouped by NAB, NASA, and YAHOO dataset groups.}
  \label{tab:ablation_full_scores}
  \centering
  \resizebox{\linewidth}{!}{%
    \begin{tabular}{lccc ccc ccc}
      \toprule
      \textbf{Variant}
        & \multicolumn{3}{c}{\textbf{NAB}}
        & \multicolumn{3}{c}{\textbf{NASA}}
        & \multicolumn{3}{c}{\textbf{YAHOO}} \\
      \cmidrule(lr){2-4}\cmidrule(lr){5-7}\cmidrule(lr){8-10}
        & \(\alpha=0.1\) & \(\alpha=0.01\) & \(\alpha=0.001\)
        & \(\alpha=0.1\) & \(\alpha=0.01\) & \(\alpha=0.001\)
        & \(\alpha=0.1\) & \(\alpha=0.01\) & \(\alpha=0.001\) \\
      \midrule
      default           & 0.384 & 0.513 & 0.450 & 0.454 & 0.610 & 0.613 & 0.623 & 0.598 & 0.492 \\
      all-pairs comparison        & 0.467 & 0.573 & 0.558 & 0.468 & 0.645 & 0.652 & 0.664 & 0.599 & 0.474 \\
      quantile $q=$12.5\%              & 0.382 & 0.506 & 0.452 & 0.451 & 0.612 & 0.599 & 0.609 & 0.592 & 0.497 \\
      quantile $q=$50\%                & 0.395 & 0.553 & 0.488 & 0.435 & 0.647 & 0.632 & 0.638 & 0.608 & 0.483 \\
      w/o class token    & 0.380 & 0.476 & 0.446 & 0.394 & 0.616 & 0.597 & 0.520 & 0.491 & 0.342 \\
      w/o cross-ref      & 0.369 & 0.478 & 0.374 & 0.416 & 0.586 & 0.587 & 0.477 & 0.436 & 0.307 \\
      w/o multi-scale    & 0.357 & 0.515 & 0.458 & 0.347 & 0.554 & 0.547 & 0.599 & 0.608 & 0.512 \\
      w/o patch-level    & 0.491 & 0.434 & 0.334 & 0.539 & 0.576 & 0.501 & 0.541 & 0.183 & 0.027 \\
      \bottomrule
    \end{tabular}%
  }
\end{table}

\begin{figure}
    \centering
    \includegraphics[width=1.0\linewidth]{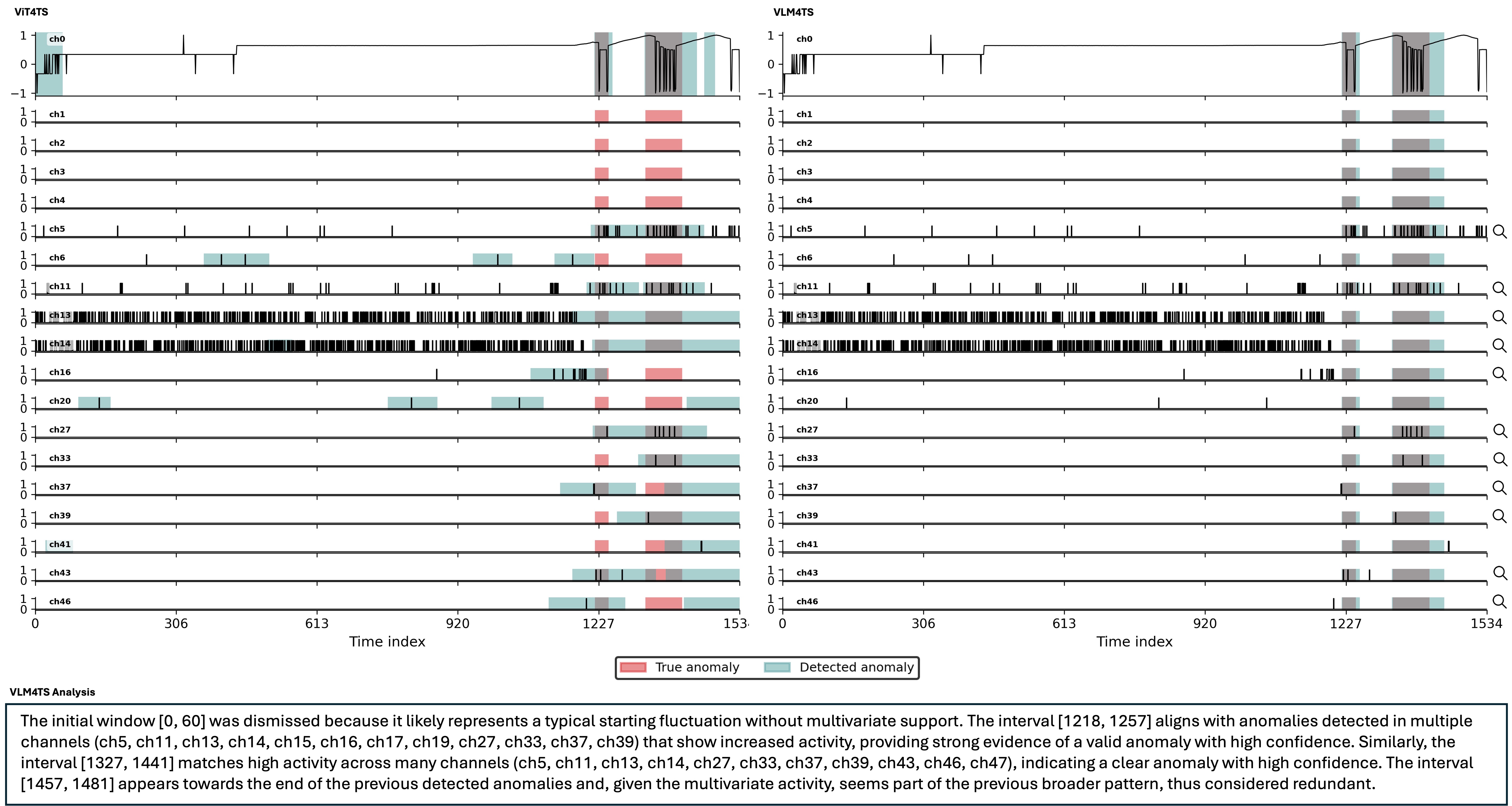}
\caption{
Multivariate extension of our framework. 
Left: ViT4TS detects candidate anomalies independently for each channel. 
Bottom: VLM4TS receives both the multivariate  visualization and a textual summary of all channel–wise candidates, enabling cross–channel visual understanding for a refined detection. 
Right: final multivariate anomaly intervals produced by VLM4TS.
}    \label{fig:placeholder}
\end{figure}

\subsection{Multivariate Extension.}

While our main experiments focus on univariate time series to highlight the key idea of the two–stage framework, the method extends naturally to multivariate settings. The key idea is to keep the first stage channel–specific and allow the second stage to perform cross–channel visual understanding and reasoning to produce the final detection.

Figure~\ref{fig:placeholder} illustrates this procedure. On the left, ViT4TS is applied independently to each channel, producing channel–wise candidate intervals based on local shape deviations. These raw candidates often include false positives or isolated fluctuations. In the second stage, VLM4TS receives (i) an aggregated textual summary of all channel–wise candidates and (ii) a stacked‐subplot visualization of the full multivariate series. As shown in the bottom panel in Figure~\ref{fig:placeholder}, with access to this global view, the VLM is able to reason across channels, retaining intervals that exhibit coordinated cross–channel anomalies while discarding channel–specific noise. The right panel shows the resulting multivariate anomaly intervals.

We conduct preliminary experiments on the multivariate MSL benchmark using this pipeline. Incorporating cross–channel context raises the Max-F1 from 0.619 (univariate) to 0.660, suggesting that VLM–based integration across channels provides additional benefit beyond single–channel counterpart. This multivariate extension preserves the design of the univariate framework while leveraging the VLM’s ability to fuse information from different time series channels.

\newtcolorbox{promptbox}{
  colback=gray!10, colframe=gray!60,
  fonttitle=\bfseries, title=VLM4TS Prompt
}

\begin{promptbox}
You are an expert in both time‐series analysis and multimodal (vision + language) reasoning. You will be shown:

1. \textbf{A plot of raw time‐series data}  
   \begin{itemize}
     \item X‐axis: time step index  
     \item Y‐axis: signal value over time  
   \end{itemize}

2. \textbf{Preliminary “vision‐based” anomaly windows}  
   \begin{itemize}
     \item A list of intervals detected by a coarse, purely visual model (may include false positives and false negatives)  
   \end{itemize}

Your goal is to \textbf{integrate both sources}—the visual plot and the preliminary windows—and produce a \textbf{refined, final anomaly detection} for the entire series. Specifically:
\begin{itemize}
  \item \textbf{Eliminate} any preliminary windows that look anomalous in isolation but are consistent with the overall trend.  
  \item \textbf{Add} any intervals that the visual model missed but which break temporal continuity or exhibit clear statistical irregularities (spikes, level shifts, abrupt changes).  
\end{itemize}

\textbf{Response format}  
Reply \textbf{only} with a JSON object containing these fields:
\begin{verbatim}
{
  "interval_index": [[start1,end1],[start2,end2],…],
  "confidence":    [c1,c2,…],
  "abnormal_description": "..."
}
\end{verbatim}
where:
\begin{itemize}
  \item \texttt{"interval\_index"}: an array of [start, end] pairs (inclusive indices).  
  \item \texttt{"confidence"}: a parallel array of integers (1–3 scale).  
  \item \texttt{"abnormal\_description"}: a single paragraph (less than 100 words) summarizing why these intervals are anomalous.  
\end{itemize}

\textbf{Confidence scale:}
\begin{itemize}
  \item 1 = Low confidence: ambiguous or very subtle deviation.  
  \item 2 = Medium confidence: clear local irregularity but moderate global uncertainty.
  \item 3 = High confidence: strong statistical or contextual evidence of anomaly.
\end{itemize}

\textbf{Important:}
\begin{itemize}
  \item Estimate interval boundaries using the tick marks on the x‐axis as precisely as possible.  
  \item The very first segment may appear atypical due to slicing; do not flag it without clear anomaly evidence.  
  \item Do not include any extra keys or commentary—only the JSON object above.  
\end{itemize}
\end{promptbox}

\subsection{Elapsed Time Comparison}
\label{app:compute}
In Table \ref{tab:compute_time}, we compare end-to-end elapsed time (measured on an NVIDIA V100 GPU) for ViT4TS, VLM4TS against both scratch-trained and time-series-pretrained TSAD baselines. We note that different class of model follows its different hyperparameter settings based on the protocol \cite{alnegheimish2024orionbench}: forecasting methods (\texttt{ARIMA}, \texttt{ATrans}, \texttt{TimesFM}, \texttt{LSTM-DT}, \texttt{UniTS}) use window size of 250 and step size of 1, and reconstruction methods (\texttt{AER}, \texttt{TadGAN}, \texttt{VAE}, \texttt{LSTM-AE}) use window size of 100 and step size of 1; ViT4TS employs a 224×224 window with a step size of 56 (one-quarter window). The reason we choose such a step size is that for each image we are applying multi-scale feature extraction and patch-level cross‐patch comparison, which allows for larger step size. For scratch-trained networks, timings include both training and inference; for pretrained models (UniTS, TimesFM) and our methods, we report inference time only. For comparison in elapsed time with language model-based methods, see Table 2. The Token usage and computation time is adjusted based on the window length and stride size, making sure that in Table 2 all the metrics are directly comparable. Despite its multi‐scale screening and VLM verification stages, ViT4TS runs in seconds on moderate‐length series and VLM4TS incurs only a few seconds per candidate via the VLM API, indicating that our two-stage pipeline is sufficiently efficient for deployment in real-world industrial monitoring systems. 
\begin{table}[ht]
  \centering
  \captionsetup{skip=4pt}
  \caption{Elapsed time comparison of VLM4TS versus trained-from-scratch and time-series-pretrained baselines on benchmark datasets. Different methods follow different hyperparameter settings, with details discussed in the Appendix.}
  \label{tab:compute_time}
  \resizebox{\linewidth}{!}{%
    \begin{tabular}{lccccccccccc}
      \toprule
      \textbf{Method}
        & \textbf{A1} & \textbf{A2} & \textbf{A3} & \textbf{A4}
        & \textbf{Art} & \textbf{AWS} & \textbf{AdEx}
        & \textbf{Traf} & \textbf{Tweets}
        & \textbf{MSL} & \textbf{SMAP} \\
      \midrule
      \texttt{ARIMA}        & 93.91  & 87.96  & 89.89  & 92.20  & 98.42   & 137.91  & 108.23  & 315.87  & 698.62  & 203.07  & 823.64  \\
      \texttt{AER}          & 20.25  & 21.70  & 24.08  & 24.55  & 37.83   & 22.90   & 22.14   & 38.09   & 57.37   & 26.48   & 33.45   \\
      \texttt{TadGAN}       & 567.06 & 549.82 & 552.48 & 561.97 & 351.49  & 358.08  & 286.88  & 752.34  & 1708.46 & 845.98  & 994.65  \\
      \texttt{LSTM-AE}      & 11.10  & 12.05  & 13.38  & 12.67  & 14.52   & 13.16   & 12.98   & 18.95   & 28.50   & 14.75   & 19.94   \\
      \texttt{ATransFormer} & 8.36   & 8.38   & 9.92   & 9.98   & 11.99   & 12.76   & 10.11   & 23.73   & 47.37   & 13.01   & 19.80   \\
      \texttt{TimesFM}      & 7.55   & 7.49   & 8.69   & 8.88   & 10.43   & 11.87   & 8.63    & 24.60   & 63.07   & 14.19   & 40.73   \\
     \texttt{TimesFM2} 
& 5.53 & 5.49 & 6.06 & 5.51 
& 6.00 & 5.97 & 5.55 
& 8.96 & 14.76 
& 7.30 & 14.27 \\
      \texttt{LSTM-DT}      & 14.61  & 12.12  & 17.34  & 17.31  & 23.06   & 15.96   & 19.01   & 33.17   & 37.45   & 20.71   & 25.10   \\
      \texttt{VAE}          & 21.66  & 24.14  & 26.59  & 25.05  & 29.05   & 26.70   & 27.17   & 70.84   & 88.90   & 30.46   & 39.17   \\
      \texttt{UniTS}        & 11.69  & 11.79  & 14.44  & 14.32  & 17.34   & 17.26   & 14.13   & 38.87   & 78.87   & 23.94   & 75.57   \\
      \midrule
      \textbf{ViT4TS}       & 2.53   & 2.26   & 3.56   & 3.58   & 7.45    & 7.23    & 3.25    & 3.74    & 44.37   & 9.56    & 25.99   \\
      \textbf{VLM4TS}       & 5.80   & 5.49   & 7.26   & 6.89   & 10.41   & 10.89   & 6.63    & 7.14    & 48.48   & 12.84   & 29.09   \\
      \bottomrule
    \end{tabular}%
  }
\end{table}

\subsection{Embedding Analysis}
\label{app:embedding}
\begin{figure}
    \centering
    \includegraphics[width=\linewidth]{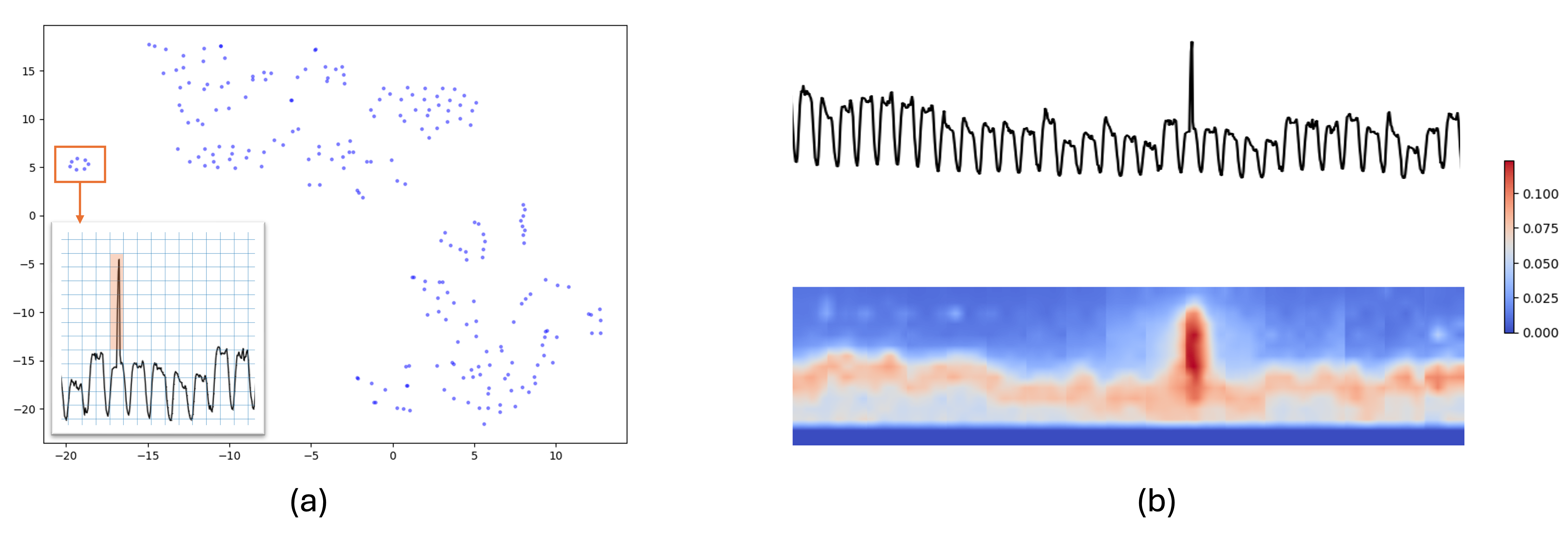}
    \caption{Patch embedding and 2-D anomaly score visualization for YAHOOA1-real\_13. Panels (a) project ViT4TS’s patch embeddings at patch scales into low-dimensional space with t-SNE, showing the clustering of anomaly-related patches (orange). Panel (b) shows the multi-scale cross-patch comparison anomaly map as a heatmap, where warmer colors indicate higher anomaly scores and identify the anomalous interval.}
    \label{fig:heatmap}
\end{figure}

Figure \ref{fig:heatmap} illustrates how ViT4TS’s patch-level embeddings distinguish anomalous regions from normal patterns. Panels (a) show a two-dimensional projection with t-SNE of the patch-level feature maps, for a representative window in A1 dataset containing the sudden spike anomaly. In panel (a), patches covering the anomaly (orange) form a compact cluster that is well separated from normal patches (blue), reflecting the pre-trained Vision Transformer’s sensitivity to local shape deviations.

Panel (b) presents the median-reference cross-patch dissimilarity heatmap over a longer series. Bright red bands precisely mark the true anomalous region, demonstrating that ViT4TS localizes anomalies in the 2-D plot before mapping back to time steps. The thin dark strip at the bottom of the heatmap corresponds to whitespace rows, which appear in every window and thus yield uniformly low dissimilarity (high similarity) under the median-reference comparison.

\subsection{Recall and Precision Comparison between ViT4TS and VLM4TS}
\label{app:recall_precision}

Tables \ref{tab:prec_rec_group1}, \ref{tab:prec_rec_group2}, and \ref{tab:prec_rec_group3} present precision and recall for ViT4TS and VLM4TS across all 11 benchmarks under three threshold settings. As expected, the verification stage usually increases precision: in most datasets, VLM4TS reduces false positives generated by the high‐recall ViT4TS screening, particularly at lower thresholds where ViT4TS produces many candidate intervals. This precision gain comes with a modest recall reduction, since VLM4TS filters out some true positives; however, the net effect is an improved F1-max in most datasets. The exceptions are A3 and A4 datasets, where the abundance of anomalies makes conservative filtering less beneficial.

On the other hand, VLM4TS also boosts recall in datasets dominated by contextual anomalies, like MSL, SMAP, Art, by flagging additional intervals that ViT4TS missed due to its limited local window. In these cases, the model’s global reasoning corrects false negatives and yields a more balanced precision–recall profile. Overall, VLM4TS improves the screening baseline’s F1-max in nine out of eleven tasks, showing the necessity of the two‐stage localization and verification design.  

\begin{table}[ht]
  \centering
  \captionsetup{skip=4pt}
  \caption{Precision and recall of ViT4TS and VLM4TS at  different thresholding quantile $\alpha$ on MSL, SMAP, and realTweets (Tweets) datasets.}
  \label{tab:prec_rec_group1}
  \resizebox{0.77\linewidth}{!}{%
    \begin{tabular}{llccc ccc ccc}
      \toprule
      \multirow{2}{*}{\textbf{Method}} & \multirow{2}{*}{\textbf{Metric}}
        & \multicolumn{3}{c}{\textbf{MSL}}
        & \multicolumn{3}{c}{\textbf{SMAP}}
        & \multicolumn{3}{c}{\textbf{Tweets}} \\
      \cmidrule(lr){3-5}\cmidrule(lr){6-8}\cmidrule(lr){9-11}
        & & 0.1    & 0.01   & 0.001
        & 0.1    & 0.01   & 0.001
        & 0.1    & 0.01   & 0.001 \\
      \midrule
      \multirow{2}{*}{ViT4TS}
        & Precision & 0.271 & 0.472 & 0.593 & 0.326 & 0.598 & 0.831 & 0.252 & 0.413 & 0.590 \\
        & Recall    & 0.978 & 0.641 & 0.432 & 0.940 & 0.779 & 0.645 & 0.921 & 0.816 & 0.605 \\
      \midrule
      \multirow{2}{*}{VLM4TS}
        & Precision & 0.404 & 0.508 & 0.397 & 0.560 & 0.731 & 0.667 & 0.727 & 0.481 & 0.360 \\
        & Recall    & 0.900 & 0.789 & 0.694 & 0.903 & 0.819 & 0.712 & 0.727 & 0.703 & 0.649 \\
      \bottomrule
    \end{tabular}%
  }
\end{table}

\begin{table}[ht]
  \centering
  \captionsetup{skip=4pt}
  \caption{Precision and recall of ViT4TS and VLM4TS at  different thresholding quantile $\alpha$ on artificialWithAnomaly ({Art}), realAWSCloudwatch (AWS), realAdExchange (AdEx), and realTraffic (Traf).}
  \label{tab:prec_rec_group2}
  \resizebox{\linewidth}{!}{%
    \begin{tabular}{llccc ccc ccc ccc}
      \toprule
      \multirow{2}{*}{\textbf{Method}} & \multirow{2}{*}{\textbf{Metric}}
        & \multicolumn{3}{c}{\textbf{Art}}
        & \multicolumn{3}{c}{\textbf{AWS}}
        & \multicolumn{3}{c}{\textbf{AdEx}}
        & \multicolumn{3}{c}{\textbf{Traf}} \\
      \cmidrule(lr){3-5}\cmidrule(lr){6-8}\cmidrule(lr){9-11}\cmidrule(lr){12-14}
        & & 0.1    & 0.01   & 0.001
        & 0.1    & 0.01   & 0.001
        & 0.1    & 0.01   & 0.001
        & 0.1    & 0.01   & 0.001 \\
      \midrule
      \multirow{2}{*}{ViT4TS}
        & Precision & 0.152 & 0.750 & 1.000 & 0.233 & 0.286 & 0.643 & 0.276 & 0.571 & 0.500 & 0.353 & 0.391 & 0.727 \\
        & Recall    & 0.909 & 0.429 & 0.167 & 0.857 & 0.516 & 0.290 & 0.727 & 0.667 & 0.273 & 0.857 & 0.643 & 0.533 \\
      \midrule
      \multirow{2}{*}{VLM4TS}
        & Precision & 0.381 & 0.546 & 0.625 & 0.267 & 0.339 & 0.400 & 0.500 & 0.615 & 0.727 & 0.500 & 0.440 & 0.526 \\
        & Recall    & 0.889 & 0.857 & 0.833 & 0.848 & 0.645 & 0.625 & 0.727 & 0.667 & 0.727 & 0.857 & 0.785 & 0.714 \\
      \bottomrule
    \end{tabular}%
  }
\end{table}

\begin{table}[ht]
  \centering
  \captionsetup{skip=4pt}
  \caption{Precision and recall of ViT4TS and VLM4TS at  different thresholding quantile $\alpha$ on  YAHOO A1-A4.}
  \label{tab:prec_rec_group3}
  \resizebox{\linewidth}{!}{%
    \begin{tabular}{llccc ccc ccc ccc}
      \toprule
      \multirow{2}{*}{\textbf{Method}} & \multirow{2}{*}{\textbf{Metric}}
        & \multicolumn{3}{c}{\textbf{A1}}
        & \multicolumn{3}{c}{\textbf{A2}}
        & \multicolumn{3}{c}{\textbf{A3}}
        & \multicolumn{3}{c}{\textbf{A4}} \\
      \cmidrule(lr){3-5}\cmidrule(lr){6-8}\cmidrule(lr){9-11}\cmidrule(lr){12-14}
        & & 0.1    & 0.01   & 0.001
        & 0.1    & 0.01   & 0.001
        & 0.1    & 0.01   & 0.001
        & 0.1    & 0.01   & 0.001 \\
      \midrule
      \multirow{2}{*}{ViT4TS}
        & Precision & 0.549 & 0.701 & 0.795 & 0.543 & 0.879 & 0.978 & 0.814 & 0.909 & 0.917 & 0.731 & 0.906 & 0.924 \\
        & Recall    & 0.697 & 0.519 & 0.348 & 0.980 & 0.905 & 0.680 & 0.493 & 0.298 & 0.200 & 0.461 & 0.302 & 0.219 \\
      \midrule
      \multirow{2}{*}{VLM4TS}
        & Precision & 0.725 & 0.742 & 0.716 & 0.853 & 0.887 & 0.828 & 0.933 & 0.931 & 0.847 & 0.900 & 0.910 & 0.836 \\
        & Recall    & 0.742 & 0.672 & 0.596 & 0.955 & 0.905 & 0.745 & 0.339 & 0.257 & 0.182 & 0.322 & 0.255 & 0.207 \\
      \bottomrule
    \end{tabular}%
  }
\end{table}

\subsection{Choice of Patch Size and Its Effect on Screening Performance}

\begin{figure}
    \centering
    \includegraphics[width=1.00\linewidth]{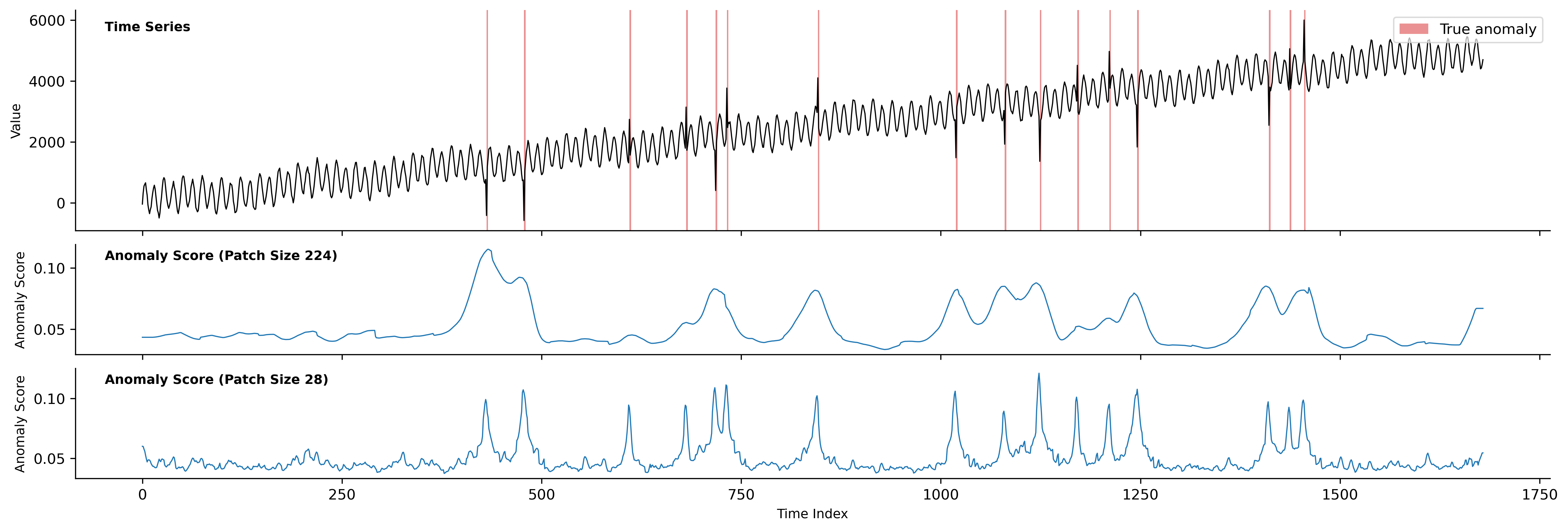}
    \caption{Patch-size effect on ViT4TS for anomaly scoring on A3 TS2. Top: original time series with ground-truth anomalies (red).
Middle: anomaly scores produced using the default patch size 224, which smooths over short and densely occurring anomalies.
Bottom: anomaly scores using a smaller patch size 28, yielding substantially improved localization.}
    \label{fig:patchsize_analysis}
\end{figure}

In Table~\ref{tab:F1versusTS} (Max-F1) and Table~\ref{tab:prec_rec_group3} (precision--recall), the ViT4TS screening stage shows noticeably lower recall on the A3 and A4 datasets, resulting in reduced Max-F1 compared with the best TSAD baselines. To understand this behavior, we analyze how ViT4TS responds to different anomaly structures. As shown in Figure~\ref{fig:patchsize_analysis}, anomalies in A3 and A4 are short, dense, and concentrated within narrow intervals. Under the default patch configuration (window size 224), several such point anomalies fall within a single window, producing coarse anomaly maps that under-represent the fine-scale deviations.

To further understand the effect of patch size, we evaluated a smaller patch size (window size 28), which offers substantially finer temporal granularity for cross-patch comparison. This adjustment yields a marked improvement, increasing Max-F1 from 0.614 to 0.906 on A3 and from 0.565 to 0.844 on A4, surpassing all baselines reported in Table~\ref{tab:F1versusTS}. Figure~\ref{fig:patchsize_analysis} also illustrates this effect. Using the default patch size (window size 224; middle panel), anomaly scores merge multiple point anomalies into broad peaks, yielding coarse localization and reduced sensitivity. In contrast, the smaller patch configuration (window size 28; bottom panel) produces sharply localized anomaly responses that align closely with
the rapid, fine-scale deviations in the signal. These results underscore the importance of patch and window size selection in capturing subtle, high-frequency anomaly patterns. Developing data-driven or adaptive strategies for patch sizing is therefore a promising direction for further improving the ViT4TS screening stage.

\subsection{Choice of VLM Backbone}
\label{app:vlm_ablation}

Table \ref{tab:vlm_maxf1_nasa_yahoo} and \ref{tab:vlm_maxf1_nab} report the Max-F1 scores of VLM4TS when using with different VLM backbones, including the close-sourced Gemini-2.0 and Claude-3.5 as well as the open-sourced Qwen-VL-2.5–72B. While these models exhibit different verification behaviors across tasks, the overall performance remains at or above state-of-the-art TSAD baselines presented in Table~\ref{tab:F1versusTS}.

\begin{table}[ht]
  \captionsetup{skip=4pt}
  \caption{Max‐F1 scores of VLM4TS with different VLMs on NASA and YAHOO dataset groups, with percentage change relative to GPT-4o.}
  \label{tab:vlm_maxf1_nasa_yahoo}
  \centering
  \resizebox{\linewidth}{!}{%
    \begin{tabular}{lcc cccc}
      \toprule
      & \multicolumn{2}{c}{\textbf{NASA}}
      & \multicolumn{4}{c}{\textbf{YAHOO}} \\
      \cmidrule(lr){2-3} \cmidrule(lr){4-7}
      \textbf{Method}
        & \textbf{MSL} & \textbf{SMAP}
        & \textbf{A1} & \textbf{A2} & \textbf{A3} & \textbf{A4} \\
      \midrule
      GPT-4o      & 
        0.619 & 0.773 
        & 0.733 & 0.901 & 0.497 & 0.474 \\

      Claude-3.5  & 
        0.588~(\(-5.0\%\)) & 0.723~(\(-6.5\%\)) 
        & 0.708~(\(-3.4\%\)) & 0.887~(\(-1.6\%\)) 
        & 0.530~(\(+6.6\%\)) & 0.490~(\(+3.4\%\)) \\

      Gemini-2.0  & 
        0.554~(\(-10.4\%\)) & 0.754~(\(-2.5\%\)) 
        & 0.723~(\(-1.4\%\)) & 0.865~(\(-4.0\%\)) 
        & 0.531~(\(+6.8\%\)) & 0.496~(\(+4.6\%\)) \\

    Qwen-2.5-72B &
  0.606~(\(-2.1\%\)) & 0.805~(\(+4.1\%\)) &
  0.747~(\(+1.9\%\)) & 0.919~(\(+2.0\%\)) &
  0.492~(\(-1.0\%\)) & 0.531~(\(+12.0\%\)) \\
      \bottomrule
    \end{tabular}%
  }
\end{table}

\begin{table}[ht]
  \captionsetup{skip=4pt}
  \caption{Max‐F1 scores of VLM4TS with different VLMs on NAB datasest group, with percentage change relative to GPT-4o.}
  \label{tab:vlm_maxf1_nab}
  \centering
  \resizebox{0.9\linewidth}{!}{%
    \begin{tabular}{lccccc}
      \toprule
      & \multicolumn{5}{c}{\textbf{NAB}} \\
      \cmidrule(lr){2-6}
      \textbf{Method}
        & \textbf{AdEx} & \textbf{Tweets} & \textbf{Traffic} & \textbf{AWS} & \textbf{Art} \\
      \midrule
      GPT-4o      & 
        0.727 & 0.686 & 0.632 & 0.488 & 0.714 \\

      Claude-3.5  & 
        0.625~(\(-14.1\%\)) & 0.617~(\(-10.0\%\)) & 0.600~(\(-4.9\%\)) & 0.543~(\(+11.2\%\)) & 0.674~(\(-5.6\%\)) \\

      Gemini-2.0  & 
        0.583~(\(-19.8\%\)) & 0.643~(\(-6.3\%\)) & 0.703~(\(+11.2\%\)) & 0.439~(\(-10.1\%\)) & 0.625~(\(-12.4\%\)) \\

    Qwen-2.5-72B &
  0.667~(\(-8.3\%\)) & 
  0.630~(\(-8.2\%\)) & 
  0.649~(\(+2.7\%\)) & 
  0.500~(\(+2.5\%\)) & 
  0.857~(\(+20.0\%\)) \\
      \bottomrule
    \end{tabular}%
  }
\end{table}

\subsection{Comparison with Spectrogram-Based TSAD Approaches.}

Spectrograms provide an alternative 2D representation of time series by encoding time–frequency structure, offering some global context. However, their use in zero-shot TSAD is restricted. Spectrogram characteristics are highly domain dependent and considerably less interpretable than line plots. Second, their visual appearance differs substantially from natural images used in large-scale vision pretraining, which weakens the transferability of off-the-shelf vision encoders and VLMs. 

It's also worth noting that spectrograms do not resolve the resolution–context dilemma: long sequences must still be divided into fixed windows before being processed, as in ITF-TAD~\citep{namura2024training}, leaving the trade-off between fine local resolution and global temporal coverage unresolved. Our two-stage design directly addresses this challenge by decoupling the tasks: ViT4TS performs fine-grained localization on high-resolution line plots, while VLM4TS conducts global verification with access to the full temporal horizon.

We compare ViT4TS and VLM4TS against the spectrogram-based ITF-TAD baseline. As shown in Table~\ref{tab:itf_tad_comparison}, ITF-TAD underperforms ViT4TS across most datasets, suggesting that spectrogram statistics are not well aligned with vision models pretrained on natural images. Furthermore, ITF-TAD falls far short of VLM4TS as it lacks a mechanism to combine fine-grained localization with global temporal understanding.

\begin{table}[h]
\centering
\caption{Comparison of ITF-TAD (spectrogram) versus ViT4TS (line-plot) and VLM4TS.}
\label{tab:itf_tad_comparison}
\resizebox{\linewidth}{!}{%
\begin{tabular}{lcccccccccccc}
\toprule
\textbf{Method} & \multicolumn{5}{c}{\textbf{NAB}} & \multicolumn{2}{c}{\textbf{NASA}} & \multicolumn{4}{c}{\textbf{YAHOO}} & \\
\cmidrule(lr){2-6}\cmidrule(lr){7-8}\cmidrule(lr){9-12}
 & \textbf{Art} & \textbf{AWS} & \textbf{AdEx} & \textbf{Traf} & \textbf{Tweets}
 & \textbf{MSL} & \textbf{SMAP}
 & \textbf{A1} & \textbf{A2} & \textbf{A3} & \textbf{A4}
 & \(\boldsymbol{\mu\pm\sigma}\) \\
\midrule

\textbf{ITF-TAD}  
  & \textbf{0.800} & 0.343 & 0.389 & 0.423 & 0.284
  & \underline{0.567} & \underline{0.726}
  & 0.569 & 0.767 & 0.425 & 0.403
  & 0.518$\pm$0.157 \\

\textbf{ViT4TS}  
  & {0.545} & \underline{0.400} & \underline{0.615} & \underline{0.615} & \underline{0.597}
  & 0.543 & \underline{0.726}
  & \underline{0.614} & \underline{0.892} & \textbf{0.614} & \textbf{0.565}
  & \underline{0.612$\pm$0.116} \\

\textbf{VLM4TS}  
  & \underline{0.714} & \textbf{0.488} & \textbf{0.727} & \textbf{0.632} & \textbf{0.686}
  & \textbf{0.619} & \textbf{0.773}
  & \textbf{0.733} & \textbf{0.901} & \underline{0.497} & \underline{0.474}
  & \textbf{0.659$\pm$0.127} \\

\bottomrule
\end{tabular}%
}
\end{table}

\subsection{Full F1 scores for Table 1}
\label{app:full_F1}

\begin{table}[ht]
  \captionsetup{skip=4pt}
  \caption{Full F1 scores of ViT4TS and VLM4TS versus trained-from-scratch and time-series-pretrained baselines for MSL, SMAP and realTweets (Tweets).}
  \label{tab:full_f1_msl_smap_tweets}
  \centering
  \resizebox{0.8\linewidth}{!}{%
    \begin{tabular}{llccc ccc ccc}
      \toprule
      \multirow{2}{*}{\textbf{Type}} & \multirow{2}{*}{\textbf{Method}}
        & \multicolumn{3}{c}{\textbf{MSL}}
        & \multicolumn{3}{c}{\textbf{SMAP}}
        & \multicolumn{3}{c}{\textbf{Tweets}} \\
      \cmidrule(lr){3-5}\cmidrule(lr){6-8}\cmidrule(lr){9-11}
        & & 0.1    & 0.01   & 0.001
        & 0.1    & 0.01   & 0.001
        & 0.1    & 0.01   & 0.001 \\
      \midrule
      \multirow{7}{*}{Trained-from-scratch}
       & \texttt{ARIMA}    & 0.409 & 0.547 & 0.585 & 0.466 & 0.750 & 0.661 & 0.094 & 0.165 & 0.179 \\
       & \texttt{AER}      & 0.475 & 0.546 & 0.553 & 0.500 & 0.711 & 0.753 & 0.130 & 0.141 & 0.178 \\
       & \texttt{TadGAN}   & 0.553 & 0.610 & 0.612 & 0.560 & 0.584 & 0.593 & 0.205 & 0.170 & 0.169 \\
       & \texttt{LSTM-DT}  & 0.592 & 0.515 & 0.615 & 0.664 & 0.724 & 0.713 & 0.076 & 0.136 & 0.190 \\
       & \texttt{LSTM-AE}  & 0.487 & 0.375 & 0.382 & 0.455 & 0.672 & 0.673 & 0.179 & 0.163 & 0.232 \\
       & \texttt{VAE}      & 0.515 & 0.342 & 0.333 & 0.553 & 0.695 & 0.661 & 0.184 & 0.156 & 0.237 \\
       & \texttt{ATrans}   & 0.387 & 0.426 & 0.454 & 0.348 & 0.538 & 0.567 & 0.147 & 0.144 & 0.145 \\
      \midrule
      \multirow{2}{*}{Time-series-pretrained}
       & \texttt{UniTS}    & 0.531 & 0.550 & 0.561 & 0.560 & 0.715 & 0.723 & 0.130 & 0.167 & 0.120 \\
       & \texttt{TimesFM}  & 0.410 & 0.500 & 0.564 & 0.434 & 0.686 & 0.602 & 0.127 & 0.182 & 0.198 \\
       & \texttt{TimesFM2} 
& 0.391 & 0.512 & 0.556 
& 0.202 & 0.608 & 0.676 
& 0.158 & 0.156 & 0.171 \\
      \midrule
      \multirow{2}{*}{Ours}
       & \textbf{ViT4TS}   & 0.425 & 0.543 & 0.500 & 0.484 & 0.677 & 0.726 & 0.395 & 0.549 & 0.597 \\
       & \textbf{VLM4TS}   & 0.558 & 0.619 & 0.505 & 0.691 & 0.773 & 0.689 & 0.486 & 0.571 & 0.686 \\
      \bottomrule
    \end{tabular}%
  }
\end{table}

\begin{table}[ht]
  \captionsetup{skip=4pt}
  \caption{Full F1 scores of ViT4TS and VLM4TS versus trained-from-scratch and time-series-pretrained baselines for {artificialWithAnomaly (Art)}, {realAWSCloudwatch (AWS)}, {realAdExchange (AdEx)}, and {realTraffic (Traf)}.}
  \label{tab:full_f1_anomaly_cloud_ad_traffic}
  \centering
  \resizebox{\linewidth}{!}{%
    \begin{tabular}{llccc ccc ccc ccc}
      \toprule
      \multirow{2}{*}{\textbf{Type}} & \multirow{2}{*}{\textbf{Method}}
        & \multicolumn{3}{c}{\textbf{Art}}   
        & \multicolumn{3}{c}{\textbf{AWS}}
        & \multicolumn{3}{c}{\textbf{AdEx}}
        & \multicolumn{3}{c}{\textbf{Traf}} \\
      \cmidrule(lr){3-5}\cmidrule(lr){6-8}\cmidrule(lr){9-11}\cmidrule(lr){12-14}
       & & 0.1   & 0.01  & 0.001  
         & 0.1   & 0.01  & 0.001  
         & 0.1   & 0.01  & 0.001  
         & 0.1   & 0.01  & 0.001  \\
      \midrule
      \multirow{8}{*}{Trained-From-Scratch}
       & \texttt{ARIMA}      & 0.344 & 0.387 & 0.353 & 0.263 & 0.185 & 0.198 & 0.295 & 0.375 & 0.500 & 0.344 & 0.320 & 0.160 \\
       & \texttt{AER}         & 0.296 & 0.318 & 0.338 & 0.244 & 0.185 & 0.185 & 0.233 & 0.412 & 0.518 & 0.404 & 0.326 & 0.323 \\
       & \texttt{TadGAN}      & 0.338 & 0.214 & 0.000 & 0.196 & 0.165 & 0.129 & 0.231 & 0.263 & 0.385 & 0.400 & 0.421 & 0.276 \\
       & \texttt{LSTM-DT}     & 0.345 & 0.368 & 0.328 & 0.273 & 0.190 & 0.202 & 0.279 & 0.367 & 0.444 & 0.451 & 0.390 & 0.414 \\
       & \texttt{LSTM-AE}     & 0.231 & 0.000 & 0.000 & 0.203 & 0.244 & 0.167 & 0.241 & 0.333 & 0.400 & 0.416 & 0.205 & 0.235 \\
       & \texttt{VAE}         & 0.000 & 0.000 & 0.000 & 0.158 & 0.248 & 0.086 & 0.200 & 0.300 & 0.345 & 0.323 & 0.190 & 0.229 \\
       & \texttt{ATrans}      & 0.239 & 0.247 & 0.262 & 0.168 & 0.140 & 0.134 & 0.200 & 0.179 & 0.157 & 0.365 & 0.324 & 0.316 \\
      \midrule
      \multirow{2}{*}{Time-series-Pretrained}
       & \texttt{UniTS}       & 0.182 & 0.000 & 0.000 & 0.228 & 0.246 & 0.214 & 0.246 & 0.326 & 0.312 & 0.479 & 0.383 & 0.323 \\
       & \texttt{TimesFM}     & 0.224 & 0.234 & 0.000 & 0.243 & 0.194 & 0.202 & 0.152 & 0.270 & 0.400 & 0.467 & 0.333 & 0.343 \\
       & \texttt{TimesFM2}
& 0.255 & 0.235 & 0.000   
& 0.233 & 0.177 & 0.176   
& 0.196 & 0.286 & 0.364   
& 0.386 & 0.381 & 0.207  \\ 
      \midrule
      \multirow{2}{*}{Ours}
       & \textbf{ViT4TS} & 0.260 & 0.545 & 0.286 & 0.366 & 0.368 & 0.400 & 0.400 & 0.615 & 0.353 & 0.500 & 0.486 & 0.615 \\
       & \textbf{VLM4TS} & 0.533 & 0.667 & 0.714 & 0.406 & 0.444 & 0.488 & 0.593 & 0.640 & 0.727 & 0.632 & 0.564 & 0.606 \\
      \bottomrule
    \end{tabular}%
  }
\end{table}

\begin{table}[ht]
  \captionsetup{skip=4pt}
  \caption{Full F1 scores of ViT4TS and VLM4TS versus trained-from-scratch and time-series-pretrained baselines for YAHOO A1-A4.}
  \label{tab:full_f1_yahoo_datasets}
  \centering
  \resizebox{\linewidth}{!}{%
    \begin{tabular}{llccc ccc ccc ccc}
      \toprule
      \multirow{2}{*}{\textbf{Type}} & \multirow{2}{*}{\textbf{Method}}
        & \multicolumn{3}{c}{\textbf{YAHOOA1}}
        & \multicolumn{3}{c}{\textbf{YAHOOA2}}
        & \multicolumn{3}{c}{\textbf{YAHOOA3}}
        & \multicolumn{3}{c}{\textbf{YAHOOA4}} \\
      \cmidrule(lr){3-5}\cmidrule(lr){6-8}\cmidrule(lr){9-11}\cmidrule(lr){12-14}
       & & 0.1    & 0.01   & 0.001
         & 0.1    & 0.01   & 0.001
         & 0.1    & 0.01   & 0.001
         & 0.1    & 0.01   & 0.001 \\
      \midrule
      \multirow{9}{*}{Trained-from-scratch}
       & \texttt{ARIMA}       & 0.269 & 0.586 & 0.650 & 0.246 & 0.663 & 0.771 & 0.502 & 0.440 & 0.179 & 0.224 & 0.336 & 0.168 \\
       & \texttt{AER}         & 0.458 & 0.614 & 0.618 & 0.232 & 0.814 & 0.866 & 0.529 & 0.696 & 0.711 & 0.399 & 0.614 & 0.609 \\
       & \texttt{TadGAN}      & 0.371 & 0.435 & 0.492 & 0.461 & 0.651 & 0.667 & 0.135 & 0.132 & 0.106 & 0.110 & 0.088 & 0.060 \\
       & \texttt{LSTM‐DT}     & 0.343 & 0.534 & 0.639 & 0.638 & 0.877 & 0.716 & 0.597 & 0.654 & 0.704 & 0.382 & 0.534 & 0.482 \\
       & \texttt{LSTM‐AE}     & 0.329 & 0.536 & 0.583 & 0.340 & 0.703 & 0.853 & 0.584 & 0.461 & 0.268 & 0.199 & 0.202 & 0.121 \\
       & \texttt{VAE}         & 0.339 & 0.537 & 0.557 & 0.343 & 0.664 & 0.845 & 0.524 & 0.353 & 0.214 & 0.189 & 0.169 & 0.090 \\
       & \texttt{ATrans}      & 0.263 & 0.239 & 0.244 & 0.481 & 0.518 & 0.554 & 0.437 & 0.334 & 0.284 & 0.394 & 0.261 & 0.212 \\
      \midrule
      \multirow{2}{*}{Time-series-pretrained}
       & \texttt{UniTS}       & 0.248 & 0.520 & 0.605 & 0.205 & 0.760 & 0.748 & 0.121 & 0.126 & 0.107 & 0.110 & 0.090 & 0.064 \\
       & \texttt{TimesFM}     & 0.221 & 0.516 & 0.554 & 0.215 & 0.694 & 0.621 & 0.120 & 0.044 & 0.015 & 0.107 & 0.066 & 0.019 \\
    & \texttt{TimesFM2}
& 0.271 & 0.577 & 0.594   
& 0.313 & 0.692 & 0.630   
& 0.202 & 0.084 & 0.013   
& 0.181 & 0.106 & 0.032   
\\
      \midrule
      \multirow{2}{*}{Ours}
       & \textbf{ViT4TS} & 0.614 & 0.597 & 0.484 & 0.699 & 0.892 & 0.802 & 0.614 & 0.449 & 0.329 & 0.565 & 0.453 & 0.354 \\
       & \textbf{VLM4TS} & 0.733 & 0.706 & 0.650 & 0.901 & 0.896 & 0.784 & 0.497 & 0.402 & 0.300 & 0.474 & 0.399 & 0.332 \\
      \bottomrule
    \end{tabular}%
  }
\end{table}

\end{document}